\documentclass{article}
\usepackage{amsmath}
\usepackage{tikz}
\usetikzlibrary{positioning, arrows.meta}
\usepackage{tabularx}
\usepackage{xltabular}
\usepackage{booktabs}
\usepackage{tcolorbox}
\tcbuselibrary{breakable,listings}
\usepackage{listings}
\usepackage{xcolor}
\usepackage{graphicx}
\usepackage{float}

\definecolor{promptbg}{RGB}{248,248,248}
\definecolor{promptborder}{RGB}{220,220,220}

\lstdefinestyle{promptstyle}{
    basicstyle=\ttfamily\small,
    breaklines=true,
    breakatwhitespace=false,
    columns=fullflexible,
    keepspaces=true,
    showstringspaces=false,
    frame=none
}

\newtcblisting{promptbox}[1]{
    breakable,
    colback=promptbg,
    colframe=promptborder,
    coltitle=white,
    colbacktitle=black,
    fonttitle=\bfseries\ttfamily,
    title=#1,
    boxrule=0.3pt,
    arc=2mm,
    left=2mm,
    right=2mm,
    top=1mm,
    bottom=1mm,
    listing only,
    listing options={style=promptstyle}
}

\PassOptionsToPackage{numbers, sort&compress}{natbib}
 \usepackage[preprint]{neurips_2026}

\usepackage[utf8]{inputenc} 
\usepackage[T1]{fontenc}    
\usepackage{hyperref}       
\usepackage{url}            
\usepackage{amsfonts}       
\usepackage{nicefrac}       
\usepackage{microtype}      

\newtheorem{remark}{Remark}
\usepackage{enumitem}

\newcommand{\bench}[1]{\texttt{CruxBench}}

\title{CruxBench: A Benchmark of Information Discovery}

\author{
 \parbox{\linewidth}{\centering
  Hui Dai\thanks{Equal contributions. This research was supported by a grant from Coefficient Giving. The views expressed in this paper do not necessarily reflect the views of the Federal Reserve Bank of Chicago or the Federal Reserve System.}\,$\;^{1}$, Lina Piao\footnotemark[1]\,$^{\;1}$, 
  Nick Merrill$^{2}$, 
  Nadja Flechner$^{2}$,
  Ezra Karger$^{2, 3}$,
  Haifeng Xu$^{1}$
}
\\
\\
$^{1}$The University of Chicago\\
$^{2}$Forecasting Research Institute \\
$^{3}$Federal Reserve Bank of Chicago\\
\small{\texttt{\{ameliadai, linap15, haifengxu\}@uchicago.edu}}
}

\begin{document}

\maketitle

\begin{abstract}
Benchmarks for large language models (LLMs) typically evaluate the accuracy of answers against fixed reference labels. But a central step in many complex real-world tasks is identifying \emph{which questions are worth asking} in the first place: decomposing a difficult problem into subquestions -- which we call \emph{cruxes} -- whose answers provide key steps on the path toward solving the target problem. To evaluate this capability of information discovery, we introduce \bench{}, a benchmark that grades LLM-generated questions by their Value of Information (VOI): how much a model-proposed crux updates beliefs about a target forecasting question. \bench{} enjoys a rare combination of three key properties: it is (1) \emph{contamination-resistant} by construction, since ground truth is generated by future world events; (2) \emph{open-ended}, admitting unbounded and complex text-based submissions rather than one correct numeric answer; and (3) \emph{grounded}, with informativeness measured against quantified changes in real-world beliefs. We evaluate a diverse set of eight models on 293 target forecasting questions and find that VOI correlates highly with independent measures of model capability ($r=0.90$) and captures cruxes' usefulness for answering target questions. However, information discovery remains challenging even for frontier LLMs, which only narrowly outperform a random-timing baseline.
\end{abstract}

\section{Introduction}
Experts and decision makers often make decisions under uncertainty, relying implicitly or explicitly on probabilistic beliefs about future outcomes, especially over long time horizons. They may confront an \emph{ultimate question} such as: Will the U.S. invade Cuba? Will there be a recession in the U.S.? And if so, over what time horizon? Answering such questions requires reasoning about a large, open-ended set of signals and intermediate events that may predict or affect the ultimate outcome. Developing accurate predictions of the ultimate outcome is therefore important, but an equally important task is to decompose the ultimate question into useful building blocks: subquestions, which we call \textit{cruxes}, whose resolution can help experts and decision makers update their beliefs about the ultimate outcome efficiently and accurately \cite{tetlock2016superforecasting, rosenberg2025belief}. Consider the following example:

\paragraph{Example.}
Consider the ultimate question: \emph{``Will there be a recession in the U.S. in the next two years?''}
An AI system might propose the following candidate cruxes:
\begin{itemize}
    \item \emph{``Will investment in AI data centers plateau or contract in the next six months?''}
    \item \emph{``Will the Democratic Party win a majority of seats in the House in the fall?''}
    \item \emph{``Will the U.S. economy show signs of weakness?''}
\end{itemize}

These cruxes differ in quality. The first corresponds to a concrete, economically relevant event whose resolution could directly and directionally update beliefs about the ultimate question. The second may also be informative, but its connection is more indirect, potentially operating through the relationship between political control and economic outcomes. The third, while intuitively related, is vague and difficult to operationalize or resolve.

\begin{figure}[ht]
\vspace{-1mm}
    \centering
    \includegraphics[width=\linewidth]{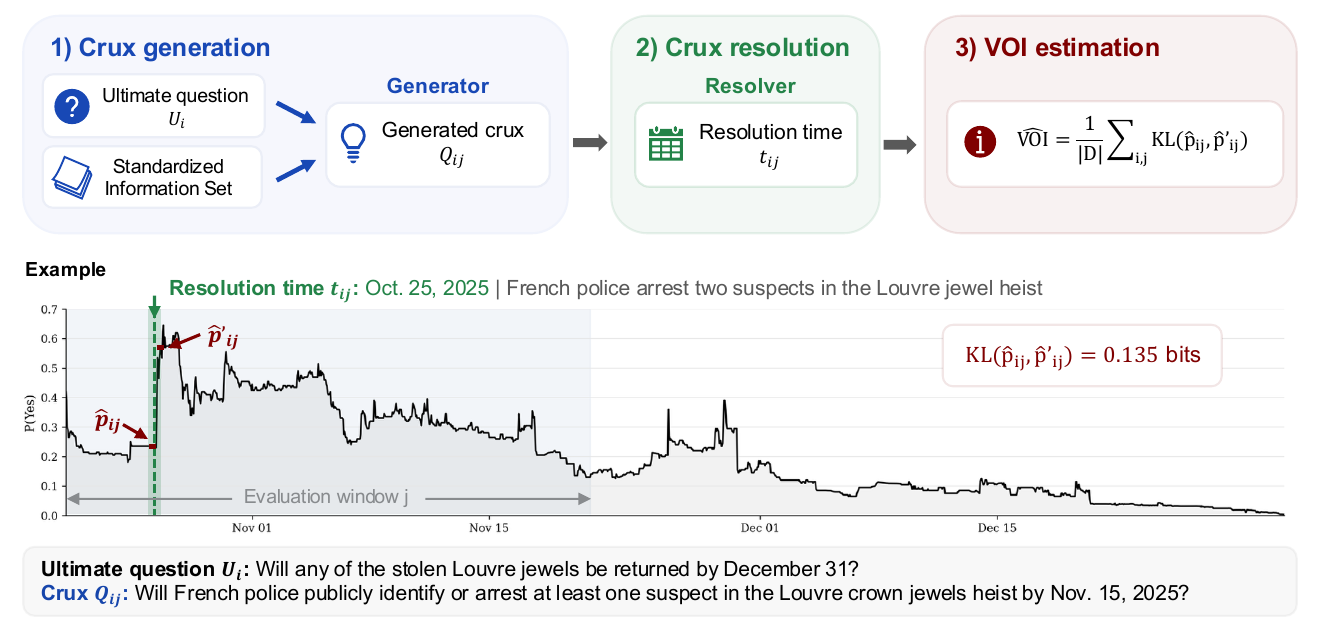}
    \caption{
    \bench{} pipeline overview. Given an ultimate question $U_i$ and the information available at the start of evaluation window $j$, a generator proposes a crux $Q_{ij}$ that is expected to resolve within the window. A separate resolver identifies the crux’s resolution time $t_{ij}$ using evidence available during the window. We measure the market-belief update around $t_{ij}$ using KL divergence and average across questions to obtain the generator’s VOI score.}
    \label{fig:pipeline}
\end{figure}

A good crux, then, is one that is specific, resolvable, and sufficiently informative that its resolution meaningfully shifts beliefs about the ultimate outcome. But how should we evaluate a model's ability to discover such cruxes? Existing evaluations of large language models (LLMs) largely overlook this capability, with most benchmarks assessing performance on fixed question--answer pairs and static datasets that are vulnerable to training-data contamination. Such benchmarks can test whether a model produces the correct answer or prediction, but they do not capture the open-ended task of proposing new, decision-relevant questions. Forecasting benchmarks address part of this limitation by grounding evaluation in future events, but they still focus primarily on predictive accuracy rather than the discovery of informative intermediate signals \cite{karger2025forecastbench, zeng2026futurex, yang2026llm}.

We therefore build on forecasting as a contamination-resistant, real-world-grounded evaluation setting while shifting the target of evaluation from \emph{answering} questions to \emph{discovering} informative ones. We introduce \bench{},\footnote{Code and data are available at \url{https://github.com/ai-prophet/cruxbench}.} a benchmark for evaluating the ability of LLMs to generate informative subquestions. Given an ultimate forecasting question, a model must propose candidate cruxes: intermediate questions whose resolution should update beliefs about the ultimate outcome. We evaluate these cruxes not through human judgment or textual plausibility, but through their \emph{value of information} (VOI) \cite{raiffa1961applied, howard1966information, frankel2019quantifying}: the extent to which their resolution updates the belief about real-world forecasting markets. We evaluate a diverse set of 8 models across 6 model families on 293 ultimate forecasting questions, and find that the ability to generate informative cruxes correlates strongly with general model capability (Pearson $r=0.90$, $p<0.01$). Moreover, cruxes from higher-VOI models help external forecasters improve their predictions of the ultimate outcome on average. However, information discovery remains challenging even for frontier models: \texttt{Claude-Opus-4.6}, \texttt{Gemini-3.1-Pro-Preview}, and \texttt{GPT-5.4} only narrowly outperform the random-timing baseline, whereas weaker models' scores are not statistically distinguishable from it. \bench{} thus provides a forward-looking and practically relevant evaluation of information discovery, with substantial headroom for future models. 

Our contributions are summarized below:
\begin{itemize}[leftmargin=*]
\item \bench{}, a contamination-resistant, open-ended, and real-world-grounded benchmark that evaluates LLMs' ability to discover informative forecasting subquestions (i.e., cruxes) using value of information.
\item A scalable pipeline for automatically generating, resolving, and evaluating cruxes using real-world belief updates from selected, highly liquid prediction markets.
\item The empirical validation that the capability of generating high-VOI cruxes correlates strongly with independent measures of model capability and improves the prediction of ultimate outcomes.
\end{itemize}

\subsection{Additional discussion on related work}
\paragraph{LLMs in forecasting.}   There is growing interest in using LLMs to forecast future events. One line of work introduces benchmarks that measure how accurately models answer forecasting questions \citep{karger2025forecastbench, zeng2026futurex, wang2025openforecast, yang2026llm}. Another develops methods to improve forecasting performance, including agentic systems \citep{alur2025aia, murphy2026agentic}, fine-tuning on the binary outcomes of resolved events \citep{halawi2024approaching, turtel2025llms, turtel2025outcome, turtel2026future, chandak2026curating}, and post-hoc calibration \citep{dai2026aligning}. Although informative cruxes may surface within a model's reasoning process, these works primarily evaluate forecasting accuracy rather than a model's ability to select informative signals.

\paragraph{Evaluating information.} Earlier work measures information by its effect on uncertainty reduction or decision quality. \citet{nelson2005finding} surveys the main families of measures for assessing the usefulness of a question: likelihood-ratio measures such as Bayesian diagnosticity \citep{good1950probability}; entropy-based measures such as expected information gain \citep{shannon1948mathematical, lindley1956measure}, which equals the expected KL divergence of the posterior from the prior \citep{kullback1951information} and coincides with VOI under the logarithmic scoring rule \citep{raiffa1961applied, howard1966information, frankel2019quantifying}; and impact, the absolute change in beliefs \citep{wells1980estimating}. These measures are not interchangeable and can rank the same questions differently \citep{nelson2005finding, frankel2019quantifying}. In the forecasting setting in particular, \citet{rosenberg2025belief} note that likelihood-ratio metrics ignore the probability that the crux itself occurs, and adopt VOI to capture both the magnitude and the probability of belief updates. Whereas they use cruxes to study disagreement on a single ultimate question (AI existential risk) and rely on human experts' estimates to compute VOI, we address a different question: benchmarking LLM-generated cruxes at scale.

\paragraph{Automatic question generation and evaluation.} A related line of work evaluates the questions LLMs ask while collaborating with humans, focusing on information gathering, proactive communication or user engagement \citep{mazzaccara2024learning, zhang2024clamber, dong2026value}. These studies concern closed-world, simulated tasks such as flight-ticket booking, whereas \bench{} targets questions about future world events, whose resolutions are unknown at generation time and whose informativeness can support real-world decision-making under uncertainty. In forecasting, recent work uses LLMs to generate question--answer pairs from news, either to study temporal generalization \citep{dai2025daily} or to construct fine-tuning data \citep{chandak2026curating, turtel2026future}. These pipelines are retrospective: given an event that has already occurred, the model reconstructs a forecasting question whose answer is known, so their evaluation centers on validity criteria such as whether a question is future-tense and resolvable. While \citet{bosse2026automating} study the automatic generation and evaluation of genuinely forward-looking forecasting questions, they also focus on validity. \bench{} pursues a different and more demanding goal: testing whether models can identify useful future information before the ultimate outcome is known, going beyond validity.

\section{Quantifying crux quality via value of information (VOI)}
\label{sec:voiest}

Our goal is to evaluate models' ability to generate informative intermediate questions, or \emph{cruxes}, whose resolutions can shift beliefs about the ultimate outcome. In this section, we describe how we use value of information to evaluate this capability at the model level.

\subsection{Notation and task setup}

We refer to each long-horizon target as an \emph{ultimate question}. Formally, for each $i = 1,\ldots,N$, the ultimate question is a binary random variable $U_i \in \{0,1\}$, 
where $U_i=1$ denotes a Yes resolution and $U_i=0$ denotes a No resolution. At generation time, the resolution of $U_i$ is unknown. For each ultimate question $U_i$, each model generates $J$ binary cruxes $Q_{ij} \in \{0,1\}$, for $j = 1,\ldots,J$, each of which should resolve before $U_i$ and be informative about its eventual resolution.

For notational simplicity, for an ultimate question $U$ and one of its generated cruxes $Q$, we define:
\begin{itemize}
  \item $p := P(U = 1)$, the prior belief about $U$ just before $Q$
        resolves;
  \item $\mathbb{P}' := P(U = 1 \mid Q)$, the random posterior belief about $U$ given $Q$, whose randomness is inherited from $Q$; 
  \item $p^{(1)} := P(U = 1 \mid Q = 1)$ and $p^{(0)} := P(U = 1 \mid Q = 0)$, 
        the two possible posteriors;
  \item $q := P(Q = 1)$, the probability that the crux $Q$ resolves Yes. 
\end{itemize}
Then $\mathbb{P}'$ takes the value $p^{(1)}$ with probability $q$ and $p^{(0)}$ with probability $1 - q$, and the law of total probability gives $p = q\, p^{(1)} + (1 - q)\, p^{(0)}$. When referring to a specific pair $(U_i, Q_{ij})$, we attach the subscript $ij$ to these quantities (e.g., $p_{ij}$, $p'_{ij}$, $\mathbb{P}'_{ij}$). 

Our data comes from a forecasting platform, where we observe the prior belief $p$ and the realized posterior $p' \in \{p^{(1)}, p^{(0)}\}$, i.e., the value taken by $\mathbb{P}'$ on the branch of $Q$ that occurred. 
The counterfactual posterior and the crux probability $q$ are unobserved. Our question is therefore: given the observed belief pairs $(p_{ij},p'_{ij})$, how can we quantify the informativeness of a generator's cruxes $Q_{ij}$?

\subsection{Value of information}
\label{sec:voi}
We assess a crux's informativeness via the \emph{value of information} (VOI), which captures how much knowing the resolution of the crux $Q$ would reduce uncertainty about the ultimate question $U$. 
Throughout the paper, we adopt entropy as the canonical measure of uncertainty, $H(x) := -x\log x - (1-x)\log(1-x)$ for $x \in [0,1]$, with logarithms taken in base 2. VOI is then formally defined as the expected \emph{entropy reduction}: 
\begin{equation}
\mathrm{VOI}(Q, U)
\;:=\; H(p) - \mathbb{E}_Q[H(\mathbb{P}')] 
\;=\; H(p) - q H(p^{(1)}) - (1-q) H(p^{(0)}).
\label{eq:voi-entropy}
\end{equation}

Equivalently, \eqref{eq:voi-entropy} is the mutual information $I(U; Q)$, a natural measure of the information shared between two random variables. 
A crux has high VOI if its possible resolutions would, on average, move beliefs about $U$ toward greater certainty. 

\paragraph{Population-level VOI.}
Ideally, we would evaluate \eqref{eq:voi-entropy} for every ultimate--crux pair $(U_i,Q_{ij})$. 
However, two quantities are not directly observable in the real-world setting. First, we observe only one realized branch of the world: either $Q = 1$ and posterior $p^{(1)}$, or $Q = 0$ and posterior $p^{(0)}$, but never both. Since per-crux VOI is an expectation over both branches, it is not identifiable from one realized history. Second, the probability $q = P(Q = 1)$ is itself unobserved and would have to be estimated, introducing additional noise. Instead, we target VOI at the population level: 
\begin{equation}
\mathrm{VOI}
\;:=\; \mathbb{E}_{(i,j)}\!\left[\mathrm{VOI}(Q_{ij}, U_i)\right] 
\;=\; \mathbb{E}_{(i,j)}\!\left[
    H(p_{ij}) - \mathbb{E}_{Q_{ij}}\!\left[H(\mathbb{P}'_{ij})\right]
\right],
\label{eq:population-voi}
\end{equation}
where the outer expectation $\mathbb{E}_{(i,j)}$ is over an ultimate--crux pair drawn at random from the population of such pairs, and the inner expectation $\mathbb{E}_{Q_{ij}}$ is over the resolution of the crux $Q_{ij}$. 

\subsection{Estimating VOI}
\label{sec:estimators}
Let $\mathcal{D}$ denote the observed dataset of ultimate--crux pairs. For each $(i,j)\in\mathcal{D}$, we observe the prior belief $p_{ij}$ just before $Q_{ij}$ resolves and the realized posterior belief $p'_{ij}$ after $Q_{ij}$ resolves. 
To estimate \eqref{eq:population-voi} from single-branch observations, we seek a per-sample function $g \colon [0,1]^2 \to \mathbb{R}$ depending only on the observable pair $(p, p')$ such that 
\[
\mathbb{E}_Q\!\left[g(p, \mathbb{P}')\right] \;=\; \mathrm{VOI}(Q, U).
\]
For any such $g$, the resulting estimator is unbiased for the population VOI in \eqref{eq:population-voi} (and consistent under standard sampling conditions, e.g., independently sampled ultimate questions): 
\[
\widehat{\mathrm{VOI}}^{g}
\;:=\; \frac{1}{|\mathcal{D}|} \sum_{(i,j)\in\mathcal{D}}
g(p_{ij}, p'_{ij}).
\]

\paragraph{Estimator 1: Change in entropy.} A straightforward choice of $g$ is the realized reduction in entropy, 
\[
\Delta H(p,p') := H(p)-H(p'),
\]
with the corresponding unbiased estimator (see a proof in Appendix~\ref{app:unbiased})
\begin{equation}
\widehat{\mathrm{VOI}}^{\Delta H}
\;:=\; \frac{1}{|\mathcal{D}|} \sum_{(i,j)\in\mathcal{D}}
\Delta H(p_{ij}, p'_{ij}).
\label{eq:voi-dh-estimator}
\end{equation}

\paragraph{Estimator 2: KL divergence.}
Another natural choice is the KL divergence of the realized posterior from the prior, viewing each as a Bernoulli distribution over $U$:\footnote{This requires $p, p' \in (0,1)$. In practice, clipping can be applied to restrict the domain of $p, p'$. } 
\begin{equation}
\mathrm{KL}(p, p')
\;:=\; \mathrm{KL}\!\left(\mathrm{Bern}(p') \,\big\|\, \mathrm{Bern}(p)\right)
\;=\; p' \log \tfrac{p'}{p} + (1-p') \log \tfrac{1-p'}{1-p}.
\label{eq:kl-def}
\end{equation}
Note that $\mathrm{KL}(p,p')$ takes its arguments in the (prior, posterior) order of $g$, the reverse of their order inside $\mathrm{KL}(\cdot\,\|\,\cdot)$. 
This choice also yields an unbiased estimator of the population VOI (see a proof in Appendix~\ref{app:unbiased}): 
\begin{equation}
\widehat{\mathrm{VOI}}^{\mathrm{KL}}
\;:=\; \frac{1}{|\mathcal{D}|} \sum_{(i,j)\in\mathcal{D}}
\mathrm{KL}(p_{ij}, p'_{ij}).
\label{eq:voi-kl-estimator}
\end{equation}

\paragraph{Extension beyond binary outcomes.}
Although \bench{} considers binary questions, our framework can extend to multi-outcome ultimate questions by replacing Bernoulli beliefs with categorical distributions and using categorical KL divergence. We provide the full formulation in Appendix~\ref{app:unbiased}.

\begin{remark}[Choice of estimators] 
Although both $\widehat{\mathrm{VOI}}^{\Delta H}$ and $\widehat{\mathrm{VOI}}^{\mathrm{KL}}$ are unbiased estimators of the same population VOI, they may differ substantially in finite samples. Throughout the paper, we use \(\widehat{\mathrm{VOI}}^{\mathrm{KL}}\) as our primary estimator for \bench{} for several reasons: it is nonnegative on every realized update, aligns with the interpretation of information as belief revision, and has a local first-order variance advantage in small- and moderate-update regimes. See more details of theoretical and empirical comparison in Appendix~\ref{app:estimatorchoice}.
\end{remark}

\section{\bench{} construction}
\label{sec:benchmark_construction}
\subsection{Pipeline overview}
Figure~\ref{fig:pipeline} summarizes our end-to-end \bench{} pipeline for evaluating a model's ability to generate informative cruxes. To evaluate information discovery from real-world data, we require a setting where belief updates $(p, p')$ can be observed over time at scale. Therefore, a natural starting point is to use \emph{prediction markets}: platforms where users bet on the outcomes of real-world events, with the contract prices reflecting an aggregated probabilistic belief about each outcome. These prices are relatively well-calibrated estimates of event probabilities.\footnote{The median Polymarket price has Brier $=0.116$ and AUC $=0.894$ on our dataset, substantially better than the strongest model baseline $P(U)$ (\texttt{Gemini-3.1-Pro-Preview}: Brier $=0.195$, AUC $=0.727$).} Each selected market question serves as one ultimate question $U_i$. Given a fixed set $\{U_i\}$, we benchmark models in three steps.
\begin{itemize}[leftmargin=*]
\item \textbf{Step 1: Crux generation.} For each ultimate question $U_i$, we provide all evaluated models with the same fixed information set and prompt each to independently generate its own set of candidate cruxes $\{Q_{ij}\}$.
\item \textbf{Step 2: Crux resolution.} Since generated cruxes are typically not themselves traded in prediction markets, we resolve them using a separate resolver agent. For each crux, the resolver assigns a binary outcome $Q_{ij}\in\{0,1\}$ and a resolution timestamp $t_{ij}$.
\item \textbf{Step 3: VOI estimation.} For each resolved crux, we read the prior $p_{ij}$ and realized posterior $p'_{ij}$ from the price history of $U_i$ shortly before and after $t_{ij}$. Aggregating $(p_{ij}, p'_{ij})$ across all of a model's cruxes using $\widehat{\mathrm{VOI}}^{\mathrm{KL}}$ (defined as ~\eqref{eq:voi-kl-estimator}) yields the model's benchmark score.
\end{itemize}

\subsection{Construction details}
\label{sec:crux_generation}
\paragraph{Crux generation.}  For each ultimate question $U_i$ resolving at time $T_i^{\text{res}}$, we partition the interval $[\max(t_0, T_i^{\text{start}}), T_i^{\text{res}} - \tau]$ into non-overlapping 30-day windows, where $T_i^{\text{start}}$ is the question's start date, $t_0$ is a fixed cutoff, and $\tau = 3$ days is a buffer so the crux resolves before $U_i$'s own end-of-life price flows. We randomly sample $J$ windows per question. For each sampled window $[t_j^{\mathrm{start}},\, t_j^{\mathrm{end}}]$, the model is prompted to generate one crux $Q_{ij}$ that resolves within that window, therefore having $J$ cruxes per ultimate question. See Appendix~\ref{app:crux-generation-prompt} for the generation prompt.

To prevent data leakage, we set $t_0=\text{September 1, 2025}$, since all evaluated models report a knowledge cutoff or release date before this (Table~\ref{tab:voi-per-model}), and disable web search during generation. Each model, instead, receives a standardized information set consisting of up to 10 relevant news articles published before the target window start time $t_j^{\mathrm{start}}$, retrieved using the AskNews API.\footnote{AskNews~(\url{https://asknews.app}) provides quality-controlled articles with publication timestamps, allowing reliable date filtering. We avoid general search engines such as Google because their date restrictions are known to be unreliable. For example, pages may be updated after publication without a corresponding change to the displayed publication date \cite{paleka2026pitfalls}.} This both supplies the recent context and ensures that all models generate cruxes from the same underlying information. However, because models may differ in training data and cutoff dates, this procedure does not fully eliminate potential information asymmetries. Our design should therefore be interpreted as controlling for, rather than completely removing, pretraining-based differences across models.

\paragraph{Crux resolution. }
Each generated crux is resolved by a separate resolver agent, distinct from the evaluated models, with access to the AskNews API. The resolver returns whether the crux is resolvable; if so, it also returns a binary outcome $Q_{ij} \in \{0, 1\}$ and a resolution timestamp $t_{ij}$ at which sufficient evidence appears. We then use a separate timestamp-verification agent to verify resolution timestamps against the underlying public evidence and align cruxes across models that resolve on the same event. Finally, we validate the resolver against human annotations and find $92.6\%$ agreement. We measure this by randomly sampling about 20 questions from each model and checking the binary outcome and resolution time through (a) manual search, (b) verification against the resolution criteria, and (c) review of the resolver's reasoning. The resolution-label accuracy is $96.9\%$, and $95.7\%$ of correctly labeled cruxes also have the correct resolution date. See Appendix~\ref{app:crux-resolution-prompt} for the resolution prompts.

\paragraph{VOI estimation.} For each resolved crux, we estimate the prior and posterior beliefs about $U_i$ from its market price history around the resolution time $t_{ij}$:
$$
\hat{p}_{ij},\, \hat{p}'_{ij} = \text{24-hour average ``Yes'' price of } U_i \text{ before / after } t_{ij}.\footnote{The 24-hour averaging window is a design choice to smooth short-term price noise. We report ablations over alternative window lengths in Appendix~\ref{app:belief-window-robustness}.}$$
The benchmark score for a model is the empirical KL estimator from Section~\ref{sec:estimators}:
\begin{equation}
\widehat{\mathrm{VOI}}^{\mathrm{KL}}
:= \frac{1}{|\mathcal{D}|} \sum_{(i,j)\in\mathcal{D}}
\mathrm{KL}(\hat{p}_{ij}, \hat{p}'_{ij}).
\end{equation}

\section{Model evaluations on \bench{}}
\label{sec:experiments}

\subsection{Dataset}
\label{sec:dataset}
We construct the benchmark from a snapshot of all Polymarket markets pulled on 2026-03-25 from its public APIs.\footnote{\url{https://gamma-api.polymarket.com/markets}, \url{https://clob.polymarket.com/prices-history}} We select resolved binary markets that are  (i) topically suited to forecasting decomposition (excluding sports, crypto, and weather markets), and (ii) liquid and uncertain enough over our evaluation period to admit nontrivial belief updates. We additionally deduplicate near-variants of the same event: a single Polymarket event often appears as several markets at different deadlines or thresholds (e.g., \emph{Fed cuts 25\,bps} vs.\ \emph{Fed cuts 50+\,bps}). For each remaining market we sample one active 30-day evaluation window uniformly at random. The resulting dataset contains $|\mathcal{D}| = 293$ ultimate questions, each paired with one window. That is, we choose $J=1$ and generate one crux per ultimate. See Appendix~\ref{app:data-filtering} for details on data filtering. For the price histories, we construct volume-aware prices that downweight raw prices with little nearby trading activity, as described in Appendix~\ref{app:vol-aware-prices}.

\subsection{Experimental setup}
\looseness=-1 We evaluate eight generators, spanning frontier and open-weight model families, on \bench{} using the pipeline in Section~\ref{sec:benchmark_construction}.\footnote{Model access and decoding parameters can be found in Appendix~\ref{app:models}.} \texttt{Claude-Sonnet-4.6} serves as the resolver for all generators, and \texttt{Claude-Sonnet-5} as the timestamp verifier. To ensure comparability across models, unresolvable cruxes are scored as $\mathrm{KL}(\hat p, \hat p') = 0$ rather than dropping them, as models could otherwise benefit from generating vague or non-resolvable questions. We also introduce two baselines. First, a \emph{random baseline}, the average $\mathrm{KL}(\hat p, \hat p')$ over all candidate hours in the active window. Equivalently, it is the expected score of a crux whose resolution time is chosen at random, capturing the average market information gain over time. Second, a \emph{hindsight oracle}, the maximum $\mathrm{KL}(\hat p, \hat p')$ over all candidate hours in the active window, representing the moment that retrospectively shows the largest information gain. Note that it serves as an upper bound rather than an attainable target, since the largest market move may reflect noise or unexpected events that a generator could not identify in advance.

\subsection{Main findings}
\bench{} separates frontier models from weaker generators. As shown in Table~\ref{tab:voi-per-model}, only the three frontier models (\texttt{Claude-Opus-4.6}, \texttt{Gemini-3.1-Pro-Preview}, and \texttt{GPT-5.4}) significantly outperform the random baseline, while the remaining models do not significantly outperform random timing (examples in Appendix~\ref{app:examples}). At the same time, even the strongest generator remains nearly an order of magnitude below the hindsight oracle ($0.129$). Although the oracle is a retrospective ceiling that no real generator may fully reach, this gap indicates substantial headroom. We say a crux is \emph{resolvable} if its outcome can be unambiguously evaluated as true or false at a specific time by an LLM resolver. Most generators achieve high resolvability rates (Table~\ref{tab:voi-per-model}), showing that they can produce well-formed, though not necessarily informative, cruxes. For typical resolvability failure modes, see Appendix~\ref{sec:unresov}.

These results suggest that crux generation is a nontrivial capability that current models do not uniformly possess. The task requires more than producing plausible and resolvable questions. Models must identify which near-term events are likely to occur and reason about how those events would update beliefs about the ultimate question. \bench{} therefore serves as a forward-looking target as model capabilities advance, measuring progress on a capability with direct value for human decision-making: surfacing informative cruxes that guide where attention should be focused. 

\begin{table}[h!]
\caption{Performance on \bench{} ($n=293$). ``Resolvable'' reports the number and percentage of cruxes with \texttt{resolvable=true}. Random baseline is the per-window mean of $\widehat{\mathrm{VOI}}^{\mathrm{KL}}$ over all hours in the active window (random-timing baseline); oracle is the per-window maximum (best-possible pivot in hindsight). Bold indicates the best-performing model. Stars denote uncorrected one-sided paired $t$-tests against the random baseline: $^{*}\,p<0.05$, $^{**}\,p<0.01$. Cutoffs marked $^{\dagger}$ use the model's public release date as a proxy where an official training-data cutoff is not published.}
\label{tab:voi-per-model}
\centering
\setlength{\tabcolsep}{6pt}
\begin{tabular}{l c c c c}
\toprule
Model & $\widehat{\mathrm{VOI}}^{\mathrm{KL}}$ (bits) & 95\% CI & \# resolved (rate) & Knowledge cutoff \\
\midrule
Claude-Opus-4.6        & \textbf{0.016}$^{**}$          & [0.009,\,0.023] & \textbf{279 (95.2\%)} & Aug 2025 \\
Gemini-3.1-Pro-Preview & 0.014$^{*}\phantom{*}$         & [0.007,\,0.020] & 274 (93.5\%)          & Jan 2025 \\
GPT-5.4                & 0.013$^{*}\phantom{*}$         & [0.007,\,0.019] & 268 (91.5\%)          & Aug 2025 \\
DeepSeek-V3.1          & 0.012$\phantom{^{**}}$         & [0.003,\,0.020] & 262 (89.4\%)          & Aug 2025$^{\dagger}$ \\
Qwen3-235B-A22B        & 0.012$\phantom{^{**}}$         & [0.005,\,0.018] & 271 (92.5\%)          & Jul 2025$^{\dagger}$ \\
Llama-3.3-70B          & 0.010$\phantom{^{**}}$         & [0.005,\,0.016] & 252 (86.0\%)          & Dec 2023 \\
Llama-3.1-8B-Instruct  & 0.010$\phantom{^{**}}$         & [0.005,\,0.015] & 271 (92.5\%)          & Dec 2023 \\
GPT-3.5-Turbo          & 0.009$\phantom{^{**}}$         & [0.005,\,0.013] & 275 (93.9\%)          & Sep 2021 \\
\midrule
Random baseline        & 0.007$\phantom{^{**}}$         & [0.006,\,0.009] & ---                   & --- \\
Oracle                 & 0.129$\phantom{^{**}}$         & [0.100,\,0.157] & ---                   & --- \\
\bottomrule
\end{tabular}
\vspace{3mm}
\vspace{-3mm}
\end{table}

\subsection{VOI reflects model capability}
\label{sec:voi-correlates-with-capability}
\begin{figure}[htbp]
    \centering
    \includegraphics[width=\linewidth]{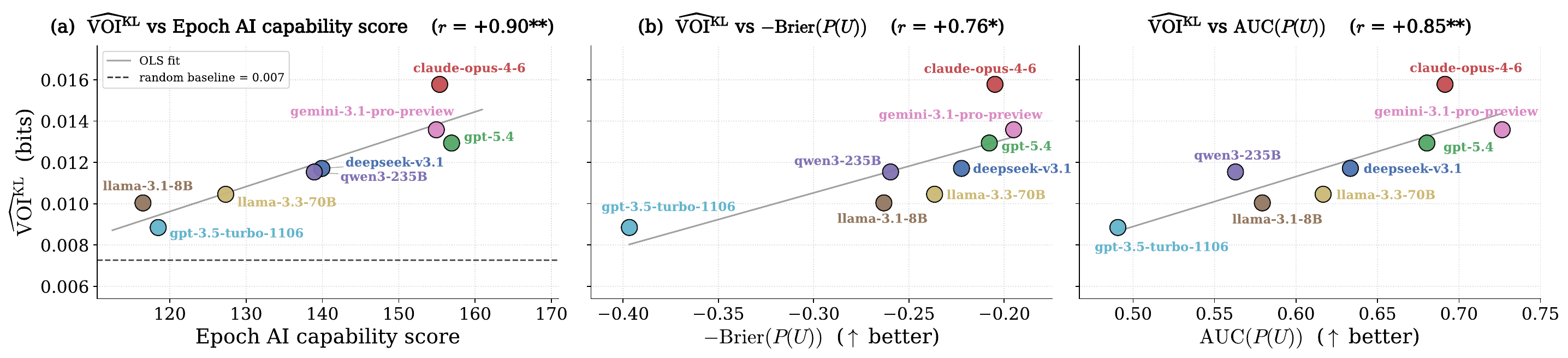}
    \caption{$\widehat{\mathrm{VOI}}^{\mathrm{KL}}$ correlates with model capability. \textbf{(a)} vs.\ Epoch AI capability score. \textbf{(b)} vs.\ each generator's forecasting performance, measured by predicting $P(U)$ at window start and scored by $-\mathrm{Brier}(P(U))$ (left) and $\mathrm{AUC}(P(U))$ (right). Brier score is sign-flipped so that higher is better.}
    \label{fig:voi_vs_capability}
\end{figure}

Figure~\ref{fig:voi_vs_capability} shows that crux quality, as measured by $\widehat{\mathrm{VOI}}^{\mathrm{KL}}$, tracks both general and forecasting-specific measures of model capability. Panel (a) compares each generator's score against the Epoch Capability Index (ECI) \citep{EpochLLMBenchmarkingHub2024}, a composite measure of general model capability that aggregates over 50 benchmarks. The two are strongly positively correlated ($r = 0.90$, $p < 0.01$). Panel (b) compares VOI against each generator's own forecasting performance on the same ultimate questions: each generator predicts $P(U)$ at window start using the information set described in Section~\ref{sec:crux_generation}, scored by negative Brier score \citep{brier1950verification} and AUC. Crux quality correlates strongly with both ($r = 0.76$ for $-\mathrm{Brier}$, $r = 0.85$ for AUC). Together, these correlations support crux generation as a capability tied to both general model performance and forecasting skill.

\subsection{Good cruxes improve downstream forecasting}
\label{sec:downstream-forecasting}
An important test of \bench{} is whether its scores predict downstream usefulness. Therefore, we examine if presented with a generator's crux, does the crux improve a forecaster's prediction of $U$, and is the improvement correlated with VOI? To test this, we evaluate a separate forecaster on each ultimate question at the window start date, with access to the same news articles available to the generator. We compare two setups: (i) a \emph{baseline} where the forecaster predicts $P(U)$ directly, and (ii) \emph{crux-informed} forecast $P(U \mid Q)$, where the forecaster predicts the ultimate given one generator's crux $Q$ and its eventual resolution. Thus, the only difference between the two is the information contributed by the resolved crux.  We also note that this setup is intentionally not realistic: in practice, a forecaster at window start date would not have selective access to one future event's outcome. We allow this leakage in order to study the contribution of a single crux's resolution to predicting $U$. Results are averaged across three external forecasters: \texttt{Claude-Haiku-4.5}, \texttt{Gemini-3-Flash-Preview}, and \texttt{GPT-5-Mini}. 

\begin{figure}[htbp]
    \centering
    \includegraphics[width=0.8\linewidth]{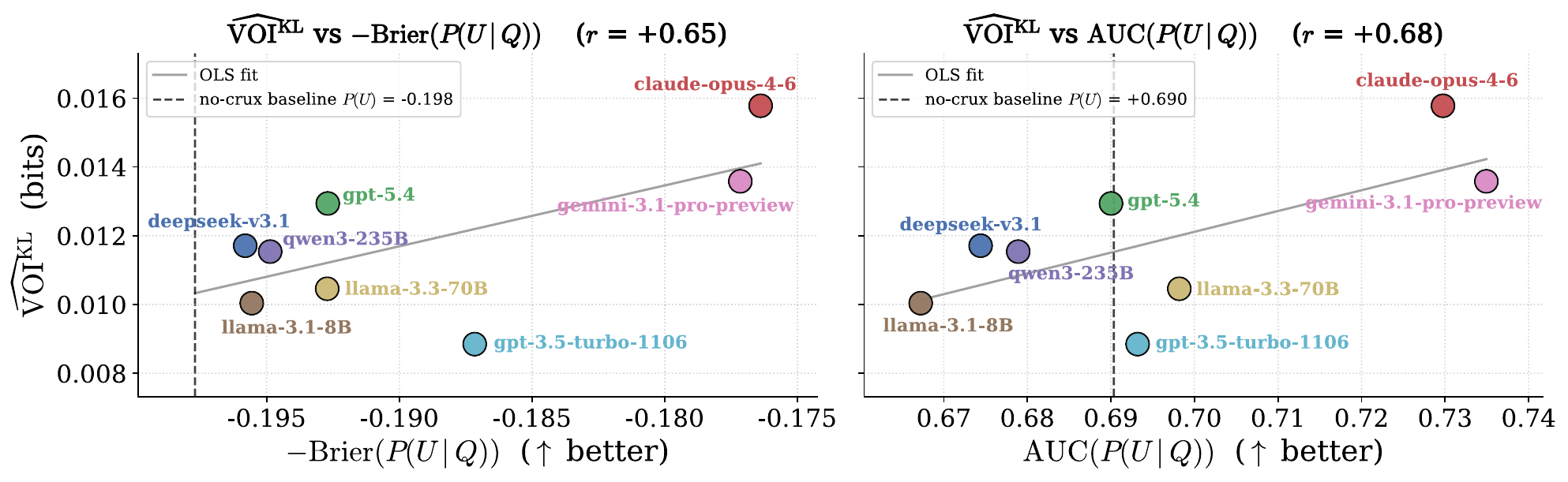} \caption{$\widehat{\mathrm{VOI}}^{\mathrm{KL}}$ correlates with downstream forecasting utility for separate forecasters. For each generator, three forecasters (\texttt{Claude-Haiku-4.5}, \texttt{Gemini-3-Flash-Preview}, \texttt{GPT-5-Mini}) predict $P(U \mid Q)$ given the crux information; results are averaged across the three.}
    \label{fig:crux_forecasting_corr}
\end{figure}

Figure~\ref{fig:crux_forecasting_corr} shows that  $\widehat{\mathrm{VOI}}^{\mathrm{KL}}$ correlates positively with downstream forecasting performance: $r=0.65$ for $-\mathrm{Brier}$ and $r=0.68$ for AUC. The highest-VOI generators, \texttt{Claude-Opus-4.6} and \texttt{Gemini-3.1-Pro-Preview}, produce the largest gains. For example, when given the resolved cruxes generated by \texttt{Claude-Opus-4.6}, the Brier score decreases from 0.198 to 0.176 and AUC increases from 0.690 to 0.730. This suggests that higher-VOI cruxes tend to be more useful to forecasters. Compared with the no-crux baseline, cruxes from all eight generators improve the Brier score, but not all improve AUC. This is plausible because small updates in the right direction can reduce squared error (better Brier score). AUC ignores probability magnitudes and depends only on ranking, so it improves only when cruxes discriminate Yes- from No-resolving ultimates.

The top two models produce the largest gains in both Brier score and AUC, whereas the third-ranked model, \texttt{GPT-5.4}, shows only a modest Brier improvement and no AUC improvement. Manual inspection suggests that this reflects a difference in crux style: many of \texttt{GPT-5.4}'s high-KL\footnote{We note that a crux’s KL is a realized belief update and should not be interpreted as that crux’s true VOI. The benchmark’s VOI scores are computed at the model level, as discussed in Section~\ref{sec:voi}.} cruxes are based on public announcements or expressions of intent rather than concrete actions or institutional commitments. Such cruxes can coincide with market movements but may provide weaker evidence about the ultimate outcome. Furthermore, VOI and downstream forecasting gains capture different aspects of crux quality: VOI measures the informativeness of a model’s generated cruxes, while downstream improvement additionally depends on whether the forecaster can effectively incorporate that crux signal. Therefore, although VOI is broadly predictive of downstream usefulness, models with similar VOI may still differ in how effectively their cruxes communicate actionable information.

\subsection{Cross-model similarity of generated cruxes} \label{sec:discussion} 

\begin{figure}[htbp]
    \centering
    \includegraphics[width=0.81\linewidth]{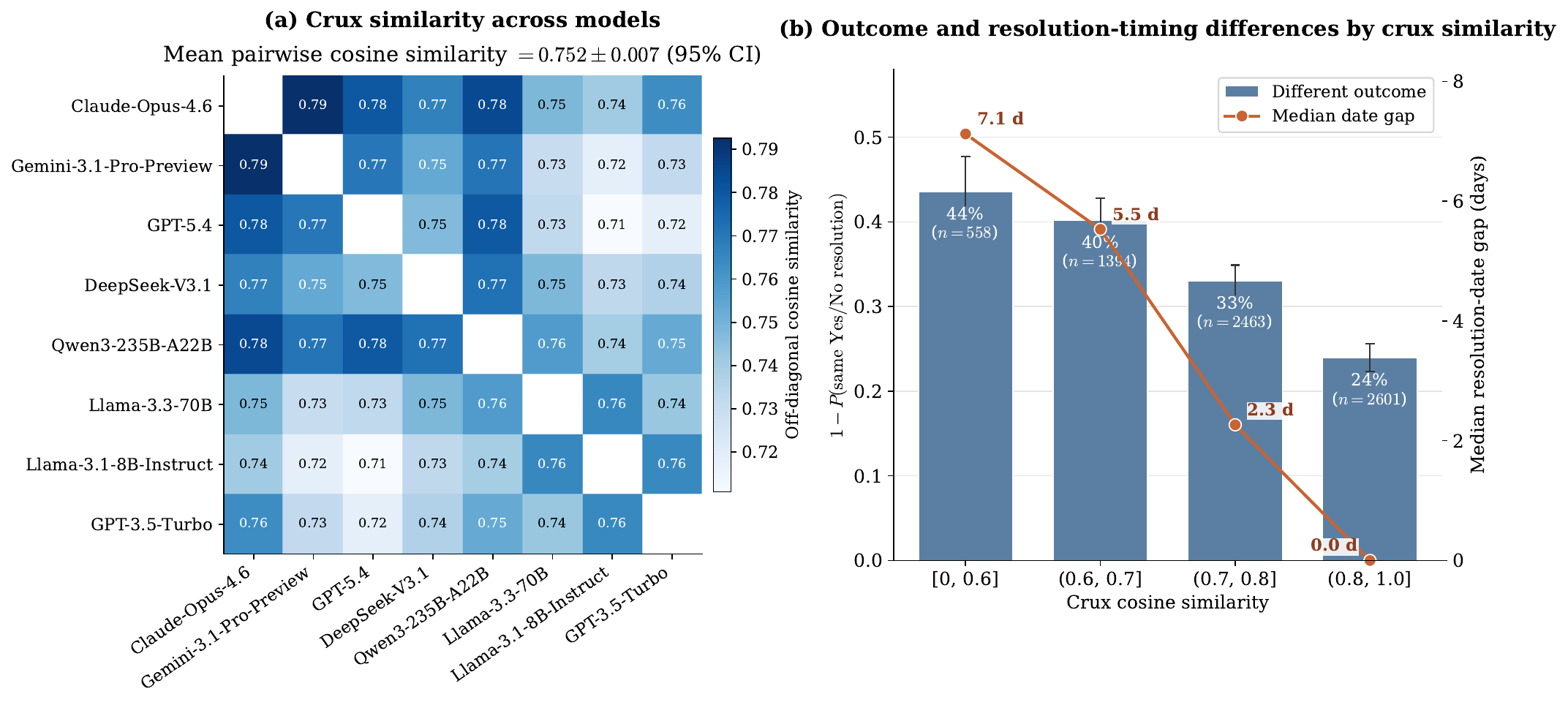}
    \caption{Semantic similarity is only a weak proxy for crux equivalence.
    \textbf{(a)} Mean cosine similarity between cruxes generated by each pair of models for the same ultimate question (mean${=}0.75$, $n=293$). \textbf{(b)} Outcome disagreement, measured as $1-P(\mathrm{same\ Yes/No\ resolution})$, and the median absolute difference in resolution dates, grouped by cosine-similarity of crux pairs.}
    \label{fig:crossmodelsimilarity}
    \vspace{-3mm}
\end{figure}

Cruxes generated by different models for the same ultimate are often semantically similar, with mean pairwise cosine similarity $\approx 0.75$ (Figure~\ref{fig:crossmodelsimilarity}(a)).\footnote{Cosine similarities are computed on the concatenated question and resolution-criteria text, embedded by OpenAI's \texttt{text-embedding-3-small} \cite{openaiembed}.} However, semantic similarity does not imply operational agreement. Among pairs with $\cos \ge 0.8$ in which both cruxes specify a numerical threshold, 84\% use different numerical thresholds (Figure~\ref{fig:sub_q_similarity_op}), and 24\% receive different Yes/No resolutions (Figure~\ref{fig:crossmodelsimilarity}(b)). These operational differences matter because thresholds and deadlines determine when a crux resolves and, consequently, whether it captures the relevant belief update.

Take the stolen Louvre jewels case in Appendix~\ref{app:examples} as an example. For the ultimate question of whether any of the jewels would be returned by December 31, \texttt{Claude-Opus-4.6} asks whether French police would identify or arrest a suspect, while \texttt{DeepSeek-V3.1} requires a suspect to be both arrested and charged. Although these two cruxes are semantically similar with $\cos=0.924$, they receive very different realized KL scores. When French police arrested two suspects on October 25, the market updated substantially, giving the first crux $\mathrm{KL}=0.135$ bits. Formal charges came four days later, after the update had largely been priced in, so the stricter crux captures only $\mathrm{KL}=0.002$ bits. Thus, even semantically similar cruxes can receive very different scores: informative crux generation requires not only identifying the right event, but also specifying an appropriate resolution criterion.

\section{Conclusion, limitations, and future directions}\label{sec:limitation}
This paper presents \bench{}, a benchmark for evaluating the ability of language models to generate informative intermediate questions for forecasting tasks. Unlike traditional benchmarks that evaluate \emph{answer} accuracy or forecasting performance, \bench{} evaluates \emph{information discovery}: the ability to identify which future events would most shift beliefs about an ultimate question. By grounding evaluation in prediction markets, \bench{} provides a scalable, open-ended, and contamination-resistant framework for studying crux generation. Across 293 forecasting questions, we find that (i) \bench{} scores align strongly with independent measures of model capability, with weaker models performing no better than a random baseline; (ii) cruxes from higher-VOI models generally improve downstream forecasting performance; and (iii) informativeness is often driven by differences in operationalization rather than semantic similarity alone. At the same time, the large gap between current models and the hindsight oracle suggests substantial room for improvement in models' ability to identify pivotal uncertainties that drive ultimate outcomes. Another open question is to discover which of $\widehat{\mathrm{VOI}}^{\Delta H}$ and $\widehat{\mathrm{VOI}}^{KL}$, or any affine combination of the two whose weights sum to one (which remains unbiased), is the most statistically efficient estimator.

 Finally, we discuss potential limitations of \bench{}, which give rise to interesting future directions. First, while per-crux VOI would be the ideal target, recovering it requires accurate estimates of the event probability $P(Q)$ and the conditional probability $P(U \mid Q)$, for which we do not have a principled estimation approach.  We therefore report only population-level estimates.  Second, we rely on market prices as a proxy for event probabilities. Although prediction markets with high liquidity are generally well-calibrated, noise remains. Moreover, observed belief updates $(p, p')$ cannot be cleanly attributed to the crux alone: there may be unrelated price fluctuations and simultaneous information arrivals contributing to the measured update. Finally, our experiments are retrospective rather than live. This design allows us to evaluate many resolved questions at scale, but it also limits our ability to test the latest models due to the data contamination concern. While our framework can be naturally extended to live evaluation, we leave it to future work.

\newpage
\begin{ack} This work is supported in part by the AI2050 program at Schmidt Sciences (Grant G-24-66104). 
\end{ack}

\bibliography{references}
\bibliographystyle{unsrtnat}

\newpage
\appendix
\section*{Appendix}
\section{Theory assumptions and proofs}
\label{app:assumptions}
\subsection{Key assumptions}

Our empirical framework identifies value of information (VOI) from observed market data under several structural assumptions. These assumptions link observed price movements to belief updates and are necessary for unbiased estimation.

\paragraph{(A1) Market prices represent beliefs.}
We assume prediction market prices correspond to probabilistic beliefs:
\[
p \approx P(U=1), \qquad p' \approx P(U=1 \mid Q).
\]

\emph{Why needed.}
Our estimators interpret $(p,p')$ as prior and posterior beliefs. Without this mapping, $\Delta H$ and KL no longer estimate VOI.

\emph{Why acceptable.}
Prediction markets are widely studied as information aggregation mechanisms and are generally well-calibrated in expectation, especially in liquid markets \cite{ng2026price}.

\paragraph{(A2) Martingale property.}
We assume posterior beliefs satisfy
\[
\mathbb{E}_Q[\mathbb{P}'] = p.
\]

\emph{Why needed.}
This condition is required for unbiasedness:
\[
\mathbb{E}_Q[\Delta H(p,\mathbb{P}')] 
= 
\mathbb{E}_Q[\mathrm{KL}(p,\mathbb{P}')] 
= 
\mathrm{VOI}(Q,U).
\]

\begin{figure}[htbp]
    \centering
    \includegraphics[width=0.5\linewidth]{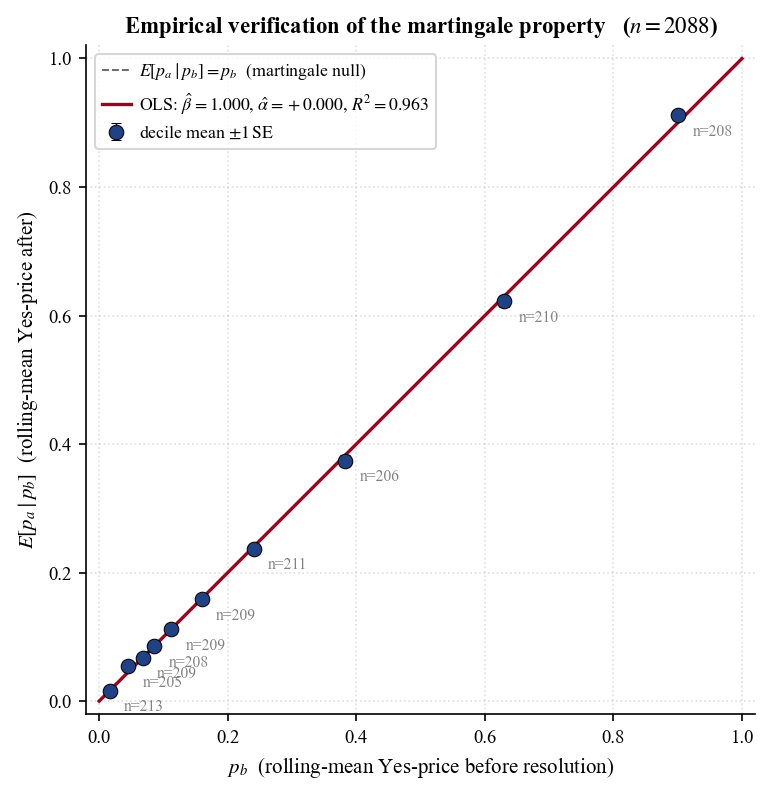}
    \caption{Empirical verification of the martingale property. Each blue point is the mean rolling-window posterior $p_a$ within a $p_b$ decile ($\pm 1\,\text{SE}$); dashed grey: 45° null $E[p_a \mid p_b] = p_b$; red: OLS fit. On $n=2{,}088$ crux-resolution pairs across 8 generators ($W=24\text{h}$), $\hat\beta = 1.00$, $\hat\alpha = +0.0002$, $R^2 = 0.96$, and decile-binned $E[p_a \mid p_b]$ lies within $1.0\text{pp}$ of the diagonal in every bin: Polymarket prices satisfy the martingale property in the mean.}
    \label{fig:martingale_check}
\end{figure}

\emph{Why acceptable.}
Under Bayesian updating with correctly specified beliefs, the prior equals the expectation of the posterior. Empirically, prediction markets approximately satisfy this calibration property, as seen below (Figure~\ref{fig:martingale_check}).

\paragraph{(A3) Attribution assumption.}
We assume that the observed belief update at 
\[
p \to p'
\]
around a crux’s resolution time $t$ reflects the informational impact of that crux.

\emph{Why needed.}
VOI is defined as the effect of observing $Q$. Without attributing the price change to $Q$, we cannot interpret $(p,p')$ as a realization of $P(U \mid Q)$.

\emph{Why acceptable.}
We focus on short windows around clearly defined events, so that the dominant information arrival is plausibly driven by the crux's resolution. While not perfectly causal, this approximation is standard in event-study settings. Furthermore, we average prices over the 24 hours before and after the resolution time, rather than relying on identification of the exact moment at which the market incorporates the information.

\paragraph{(A4) Representative branch realization.}
We observe only one realized posterior $p' \in \{p_1,p_0\}$. This realization is assumed to be drawn according to the true probability $q = P(Q=1)$.

\emph{Why needed.}
Unbiased estimation requires:
\[
\mathbb{E}_Q[g(p,\mathbb{P}')] = \mathrm{VOI}(Q,U).
\]
Selection bias in which branch is observed would violate this equality.

\emph{Why acceptable.}
Crux resolution is determined by real-world events rather than researcher choice, making realized branches plausibly representative.

\paragraph{(A5) Population-level exchangeability.}
We assume the collection of $(i,j)$ pairs has sufficient independence such that
\[
\frac{1}{|\mathcal{D}|} \sum g(p_{ij},p'_{ij})
\;\to\;
\mathbb{E}[\mathrm{VOI}(Q,U)].
\]

\emph{Why needed.}
Per-crux VOI is not identifiable; aggregation is required for consistency.

\emph{Why acceptable.}
Our dataset spans many independent forecasting questions, and VOI is normalized via entropy, making cross-question comparisons meaningful.

\medskip

These assumptions allow us to interpret observed price changes as noisy realizations of belief updates induced by crux resolution, enabling estimation of population-level VOI.

\subsection{Practical constraints}

Our empirical setup relies on several assumptions that may not hold perfectly in practice. These issues mainly affect how cleanly we can identify and measure VOI, potentially introducing bias or additional variance into our estimators. 

\paragraph{Market inefficiencies.}
Prediction market prices may not perfectly reflect true beliefs due to liquidity constraints, strategic trading, or risk premium \cite{wolfers2004prediction}. In these cases, $p$ may deviate from the true probability $P(U=1)$, weakening the interpretation of price updates as belief updates and potentially biasing VOI estimates.

\paragraph{Confounding information arrivals.}
Observed price movements may reflect multiple pieces of information arriving simultaneously rather than only the resolution of the crux. As a result, the observed update $(p,p')$ may capture the combined effect of several events instead of the isolated contribution of $Q$.

\paragraph{Timing misalignment.}
Information may be incorporated into market prices before or after our measurement window. If markets react early, the measured update is attenuated because part of the information has already been priced in. If markets react late, unrelated future information may contaminate $(p,p')$.

\paragraph{Non-event resolutions and weak signals.}
Many cruxes (64.2\%) resolve to ``No'' simply because the specified event never occurs within the assigned time window. In these cases, the resolution corresponds to the expiration of the window rather than a discrete observable event. Although the absence of an event can still be informative, these resolutions often induce only small belief updates, making $(p,p')$ more sensitive to background market noise and unrelated fluctuations.

\paragraph{Resolution errors.}
Our automated resolver pipeline may introduce timing or classification errors when using external information sources. These errors effectively add measurement noise to $(p,p')$, increasing estimator variance and potentially introducing bias if the errors are systematic.

\medskip

We partially mitigate these threats through aggregation across a large set of cruxes, and we provide the following robustness checks. 

Threat 1: the prior price satisfies the martingale property $\mathbb{E}[p' \mid p] = p$ across t-test, regression, and decile-bias checks (Figure~\ref{fig:martingale_check}). Threat 2 is addressed indirectly by the cross-model crux convergence in Section~\ref{sec:discussion}. Threat 3: model rankings are relatively stable when the 24-hour belief window is varied over $W \in \{1, \ldots, 48\}$ hours, except at smaller $W$s (Appendix~\ref{app:belief-window-robustness}). Threat 4: we impute $\mathrm{KL} = 0$ for unresolvable cruxes (Table~\ref{tab:voi-per-model}) to avoid selection bias toward easily resolvable questions (Appendix~\ref{sec:unresov}). Threat 5: hand-validation against the LLM resolver gives $92.6\%$ overall agreement, $96.9\%$ resolution-label accuracy, and $95.7\%$ timestamp agreement among correctly labeled cruxes (Section~\ref{sec:crux_generation}).

\section{Extension of the metrics and evaluations to non-binary outcomes}
\label{app:unbiased}

Suppose now that the ultimate question takes values in $\{1,\dots,K\}$ with $K \ge 2$, and that the crux takes values in a finite set $\mathcal{Q}$ (in the main text, $\mathcal{Q} = \{0,1\}$). Let $\pi = (\pi_1,\dots,\pi_K)$ denote the prior distribution over outcomes just before $Q$ resolves, where $\pi_k := P(U = k)$ and $\sum_{k=1}^K \pi_k = 1$. After the crux resolves, beliefs update to the random posterior distribution $\Pi' = (\Pi'_1,\dots,\Pi'_K)$ with $\Pi'_k := P(U = k \mid Q)$, the analogue of $\mathbb{P}'$ in the main text. For each possible crux realization $a \in \mathcal{Q}$, define
\[
\pi^{(a)}_k := P(U = k \mid Q = a)
\qquad\text{and}\qquad
q_a := P(Q = a),
\]
so that $\Pi' = \pi^{(a)}$ with probability $q_a$. By the law of total probability,
\begin{equation}
\pi = \sum_{a \in \mathcal{Q}} q_a\, \pi^{(a)},
\qquad\text{i.e.,}\qquad
\mathbb{E}_Q[\Pi'_k] = \pi_k \quad \text{for every } k.
\label{eq:ltp-cat}
\end{equation}
The entropy of a categorical distribution $\pi$ is $H(\pi) := -\sum_{k=1}^K \pi_k \log \pi_k$, with the convention $0 \log 0 = 0$; for $K = 2$ it reduces to the binary entropy $H(\pi_1)$ of the main text. The entropy-based value of information is then
\[
\mathrm{VOI}(Q,U)
\;:=\; H(\pi) - \mathbb{E}_Q[H(\Pi')]
\;=\; H(\pi) - \sum_{a \in \mathcal{Q}} q_a H(\pi^{(a)}).
\]
Equivalently, $\mathrm{VOI}(Q,U) = I(U;Q)$, the mutual information between the ultimate question and the crux.

Let $(\pi,\pi')$ be an observed prior--posterior pair, where $\pi'$ is the realized value of $\Pi'$. The two per-sample functions of the main text extend as follows: the entropy reduction is $\Delta H(\pi,\pi') := H(\pi) - H(\pi')$, 
and the natural analogue of the binary KL function is the categorical KL divergence, with arguments in the (prior, posterior) order of the main text:
\[
\mathrm{KL}(\pi,\pi')
\;:=\; \mathrm{KL}\!\left(\pi' \,\big\|\, \pi\right)
\;=\; \sum_{k=1}^K \pi'_k \log \frac{\pi'_k}{\pi_k},
\]
with the additional convention $0 \log \tfrac{0}{0} = 0$. Since $\pi_k = 0$ implies $\Pi'_k = 0$ by \eqref{eq:ltp-cat}, outcomes with zero prior probability contribute nothing to $H$, $\Delta H$ or $\mathrm{KL}$, and $\mathrm{KL}(\pi,\Pi')$ is always finite. 

\paragraph{Unbiasedness.} 
Both choices are unbiased for VOI. For the entropy reduction, this is immediate from the definition of VOI:
\begin{equation}
\mathbb{E}_Q\!\left[\Delta H(\pi,\Pi')\right]
\;=\; H(\pi) - \mathbb{E}_Q[H(\Pi')]
\;=\; \mathrm{VOI}(Q,U).
\label{eq:dh-unbiased}
\end{equation}
For the KL divergence, since $\mathrm{KL}(\pi,\pi') = -H(\pi') - \sum_{k} \pi'_k \log \pi_k$ and $H(\pi) = -\sum_{k} \pi_k \log \pi_k$, we have the pointwise identity
\begin{equation}
\mathrm{KL}(\pi,\pi')
\;=\; \Delta H(\pi,\pi') \;-\; \sum_{k=1}^K (\pi'_k - \pi_k) \log \pi_k .
\label{eq:kl-dh-identity}
\end{equation}
The last term has zero mean under $Q$, because $\mathbb{E}_Q[\Pi'_k] = \pi_k$ by \eqref{eq:ltp-cat}. Hence
\[
\mathbb{E}_Q\!\left[\mathrm{KL}(\pi,\Pi')\right]
\;=\; \mathbb{E}_Q\!\left[\Delta H(\pi,\Pi')\right]
\;=\; \mathrm{VOI}(Q,U).
\]

Therefore, averaging over observed prior--posterior pairs $(\pi_{ij},\pi'_{ij})$ yields the estimators of the population VOI 
\begin{align*}
\widehat{\mathrm{VOI}}^{\Delta H}
&:= \frac{1}{|\mathcal{D}|} \sum_{(i,j)\in\mathcal{D}} \Delta H(\pi_{ij},\pi'_{ij}), \\
\widehat{\mathrm{VOI}}^{\mathrm{KL}}
&:= \frac{1}{|\mathcal{D}|} \sum_{(i,j)\in\mathcal{D}} \mathrm{KL}(\pi_{ij},\pi'_{ij})
\;=\; \frac{1}{|\mathcal{D}|} \sum_{(i,j)\in\mathcal{D}} \sum_{k=1}^{K}
\pi'_{ij,k} \log \frac{\pi'_{ij,k}}{\pi_{ij,k}}.
\end{align*}
Here we assume, without loss of generality, that all ultimate questions have the same number of outcomes $K$: if $U_i$ has fewer than $K$ outcomes, we pad it with artificial outcomes that receive probability zero under both the prior and the posterior, which by the conventions above changes neither $H$ nor $\mathrm{KL}$. 
For $K = 2$ and $\mathcal{Q} = \{0,1\}$, identifying the Yes resolution with outcome $k = 1$ gives $\pi = (p, 1-p)$, $\pi' = (p', 1-p')$, $\pi^{(a)} = (p^{(a)}, 1-p^{(a)})$ and $q_1 = q$. This recovers the estimators \eqref{eq:voi-dh-estimator} and \eqref{eq:voi-kl-estimator} of the main text, together with their unbiasedness.

The same argument extends to continuous outcomes by replacing sums over outcomes with integrals: $\mathrm{VOI}(Q,U) = I(U;Q) = \mathbb{E}_Q[\mathrm{KL}(\Pi' \,\|\, \pi)]$, with the KL divergence taken between densities, so the KL estimator remains unbiased. The entropy form continues to hold with differential entropies whenever they are finite.

\section{Estimator choice: details}
\label{app:estimatorchoice}
\paragraph{KL is nonnegative and does not penalize movement toward uncertainty.}
A basic property of VOI is nonnegativity: observing additional information should not reduce the expected value of the decision problem. The KL estimator preserves
this property at the realized-sample level: $\mathrm{KL}(p,p') \geq 0.$ By contrast, realized entropy reduction can be negative: $\Delta H(p,p')<0$ if $H(p')>H(p).$ This happens when the posterior moves closer to the uncertainty $1/2$, even though the update may still be informative. For example, suppose we are initially confident that candidate A will win an election (prior $p = 0.9$), but a scandal breaks and we revise our belief to $p' = 0.5$. Then $\Delta H = -0.531$ while $\mathrm{KL} = 0.737$. The update is clearly informative, yet $\Delta H$ assigns it negative value. More generally, $\Delta H$ only rewards movement toward certainty, regardless of direction, whereas KL captures information content whether the update increases or decreases entropy. The latter is a closer match to the intuitive notion of crux usefulness.

\paragraph{Relationship and variance.}
The entropy-reduction and KL estimators are related by the exact identity
\begin{equation}
    \Delta H(p,p')
=
\operatorname{KL}(p,p')
+
\log\tfrac{p}{1-p}(p'-p).
\end{equation}
Thus, entropy reduction differs from KL by an additional linear term in the belief update $(p'-p)$. Since this residual term has mean zero across branches, both estimators remain unbiased for VOI. However, it contributes additional finite-sample variance.

Writing $\delta = p'-p$, a Taylor expansion gives
\begin{equation}
    \operatorname{KL}(p,p+\delta)
=
\frac{\delta^2}{2\ln (2)p(1-p)}
+
O(\delta^3),
\end{equation}
so KL begins at second order in the update size, whereas $\Delta H$ contains an extra first-order component proportional to
\[
\log\tfrac{p}{1-p}\,\delta.
\]
Consequently,
\begin{equation}
    \mathrm{Var}[\Delta H(p,p')]
-
\mathrm{Var}[\mathrm{KL}(p,p')]
=
\left[\log\tfrac{p}{1-p}\right]^2 \sigma_\delta^2
+
O\!\left(\mathbb{E}[|\delta|^3]\right),
\end{equation}
where $\sigma_\delta^2 = \mathrm{Var}(\delta)$. The leading term is nonnegative, implying that KL has lower variance to leading order whenever posterior updates are sufficiently small. Intuitively, KL acts as a variance-reduced version of entropy reduction by removing the directional first-order sensitivity of the update.

\paragraph{Estimator choice and variance simulation}
To compare the statistical efficiency of VOI estimators, we conduct a Monte Carlo study in which we simulate belief updates under a partially informative signal. In each setting, the signal reveals the true outcome with probability $\alpha$ and otherwise reflects unrelated noise, allowing us to vary informativeness from uninformative to fully revealing. Across these regimes, we compute VOI using both KL-divergence and entropy-difference formulations and compare their sampling variability across replications. This design isolates how estimator variance depends on signal informativeness and the underlying distribution of beliefs (see Appendix~\ref{app:monte-carlo} for details).

Across 121 simulation cells spanning different belief distributions and levels of informativeness $\alpha$, we find that the KL estimator has lower variance in the majority of regimes (78 out of 121 cells, 64.5\%), with a geometric mean variance ratio $\mathrm{Var}(H)/\mathrm{Var}(KL) = 1.247$, indicating that KL is more variance-efficient on average. This advantage is most pronounced in low- and moderate-informativeness regimes (small $\alpha$), where posterior updates are limited and signals are weakly informative (as seen in Figure~\ref{fig:montecarlosim} of Appendix~\ref{app:monte-carlo}). 

Hence, we report $\widehat{\mathrm{VOI}}^{\mathrm{KL}}$ as our headline benchmark metric, as it is the lower-variance estimator in the majority of regimes and especially in the low-information settings most relevant to our application (see Appendix ~\ref{app:alpha-estimation} for more details). For completeness, Appendix~\ref{app:additional-metrics} also reports results for $\widehat{\mathrm{VOI}}^{\Delta H}$, along with two additional belief-update metrics: a variance-based VOI metric $(p' - p)^2$ and the absolute update $|p' - p|$. Empirically, the latter two produce rankings broadly consistent with the KL-based measure (as is expected by the decomposition in Appendix ~\ref{app:additional-metrics}).

\section{Dataset construction}

\subsection{Data filtering}
\label{app:data-filtering}
Here we describe the filter from raw Polymarket dump to the final benchmark set summarized in Section \ref{sec:dataset}. Table~\ref{tab:dataset-funnel} shows the counts after filtering for each stage.

\paragraph{Resolved binary markets.} We retain only resolved markets with binary ``Yes'' or ``No'' options. We deduplicate markets by question text and require them to overlap with the evaluation period ($\geq t_0 =$ 2025-09-01).

\paragraph{Topic filters.}
We first exclude markets tagged as sports, crypto, or weather. We then apply keyword filters to remove remaining sports-, crypto-, and awards-related markets. These domains are often highly volatile or driven by short-term information that may be difficult to decompose into informative cruxes. We also drop questions whose text contains \texttt{tweet} or \texttt{say} (e.g., \emph{Will Trump say ``Cat'' in May 2026?}), which are usually not substantive events and contain high uncertainty. We further drop the ``any of the listed'' type of markets, because we find that the listed resolution criteria are not well defined. We also drop a small set of markets requiring access to some AI-leaderboards, whose information is hard for the resolver to gather.

\paragraph{Lifetime and liquidity filters.}
We require that the market lasts $\texttt{endDate}-\texttt{startDate} \geq 60$ days. For liquidity, we require at least 10,000 USDC in cumulative trading volume at fetch time and price histories spanning at least 30 distinct days. After randomly selecting one evaluation window per market, we further require more than 1,000 USDC in observed trading volume within that window.

\paragraph{Event-level deduplication.}
For the same events, Polymarket often has several markets around it (at different resolution dates or different numerical thresholds). To ensure diversity in our dataset, we randomly select one single market per Polymarket event ID. We also remove near-duplicates by date- and number-stripped normalization of the question text.

\paragraph{Active-window selection.}
We drop windows where the market price is within $0.1$ of $0$ or $1$ through the window (i.e., the event has effectively already resolved), and residual windows shorter than 7 days. For the final evaluation, we sample one window per ultimate question, so $J=1$.

\begin{table}[h]
\caption{Per-stage retention markets from the raw Polymarket dump to the final dataset ($|\mathcal{D}| = 293$).}
\label{tab:dataset-funnel}
\centering
\small
\begin{tabular}{lr}
\toprule
\textbf{Stage} & \textbf{Markets} \\
\midrule
Raw Polymarket dump (2026-03-25)            & 108{,}101 \\
\quad price-history end $\geq$ 2025-09-01    & 87{,}124 \\
\quad has resolution                         & 86{,}819 \\
\quad dedup by question text                 & 84{,}043 \\
\quad UMA status = resolved                  & 84{,}035 \\
\quad has \texttt{Yes} outcome               & 30{,}524 \\
\quad excluded tags (sports/crypto/weather)  & 25{,}179 \\
\quad excluded words \texttt{(tweet, say)}   & 23{,}111 \\
\quad lifetime $\geq$ 60 days                & 3{,}009 \\
\quad sports/crypto/awards keyword filter    & 1{,}414 \\
\quad price-history length $\geq$ 30         & 1{,}063 \\
\quad event-ID deduplication                 & 519 \\
\quad normalized-question deduplication      & 505 \\
\quad description-phrase blocklist           & 492 \\
\quad has $\geq 1$ valid 30-day window       & 342 \\
\quad active window $\geq$ 7 days            & 336 \\
\quad selected-window volume $>$ 1,000 USDC & \textbf{293}\\
\bottomrule
\end{tabular}
\end{table}

\subsection{Volume-aware price history construction}
\label{app:vol-aware-prices}
Polymarket's reported price history tracks the midpoint between the best bid and ask quotes in the order book. Consequently, the reported price can change when traders update or cancel their quotes, even if no trade happens. To mitigate the influence of price movements unsupported by trades, we reconstruct the price history by assigning more weight to prices with more nearby trading volume.

Let $p_t^{\mathrm{raw}}$ denote the raw hourly Yes price, and let $V_t$ be the observed USDC trading volume within a $\pm 12$-hour neighborhood of hour $t$. We define
$$
\alpha_t=\min\left(1,\frac{V_t}{1000}\right),
\qquad
p_t
=
\alpha_t p_t^{\mathrm{raw}}
+
(1-\alpha_t)p_{t-1},
$$
with $p_0=p_0^{\mathrm{raw}}$. Therefore, when nearby trading volume reaches 1,000 USDC, we trust the raw price. When no nearby trading is observed, we carry forward the previous adjusted price. For volumes between 0 and 1,000 USDC, $p_t$ is a weighted average of the raw price and the previous adjusted price, with more weight placed on the raw price as volume increases. All metrics are computed using the volume-aware price history $p_t$.

\section{Examples of cruxes}
\label{app:examples}

\subsection{Example 1: Stolen Louvre jewels}
\textbf{Ultimate question:} ``Will any of the stolen Louvre jewels be returned by December 31?''

\textbf{Evaluation window:} 2025-10-21 to 2025-11-20

\begin{table*}[htbp]
\caption{Cruxes generated for ``Will any of the stolen Louvre jewels be returned by December 31?'' with their realized $\mathrm{KL}(\hat p, \hat p')$.}
\label{tab:example_1}
\centering
\small
\begin{tabular}{p{3cm} p{9.8cm} p{1.2cm}}
\toprule
\textbf{Model} & \textbf{Proposed Crux} & \textbf{KL} \\
\midrule

Gemini-3.1-Pro-Preview &
Will authorities announce the arrest or formal charging of at least one suspect in connection with the October 2025 Louvre jewel heist between Oct.\ 21 and Nov.\ 20, 2025? &
0.1348 \\

Claude-Opus-4.6 &
Will French police publicly identify or arrest at least one suspect in the Louvre crown jewels heist by Nov.\ 15, 2025? &
0.1348 \\

GPT-5.4 &
Will French authorities publicly announce at least one arrest or formal charging of a suspect in connection with the October 19, 2025 Louvre crown-jewels theft by 2025-11-20? &
0.1348 \\

Llama-3.3-70B &
Will the French authorities announce the recovery of at least one stolen Louvre jewel or the arrest of at least one suspect in connection with the heist between 2025-10-21 and 2025-11-20? &
0.1348 \\

Qwen3-235B-A22B &
Will French law enforcement announce the arrest of at least one suspect in connection with the Louvre crown jewels heist between Oct.\ 21, 2025, and Nov.\ 20, 2025? &
0.1348 \\

DeepSeek-V3.1 &
Will at least one suspect involved in the Louvre crown jewels heist be formally arrested and charged by Nov.\ 15, 2025? &
0.0022 \\

Llama-3.1-8B-Instruct &
Will the French authorities have recovered at least one of the stolen Louvre jewels by Nov.\ 15, 2025? &
0.0002 \\

GPT-3.5-Turbo &
Will any of the stolen Louvre jewels appear on the public market by Nov.\ 20, 2025? (This includes online listings or auction announcements.) &
0.0002 \\

\bottomrule
\end{tabular}
\end{table*}

\begin{figure}[htbp]
    \centering
    \includegraphics[width=\linewidth]{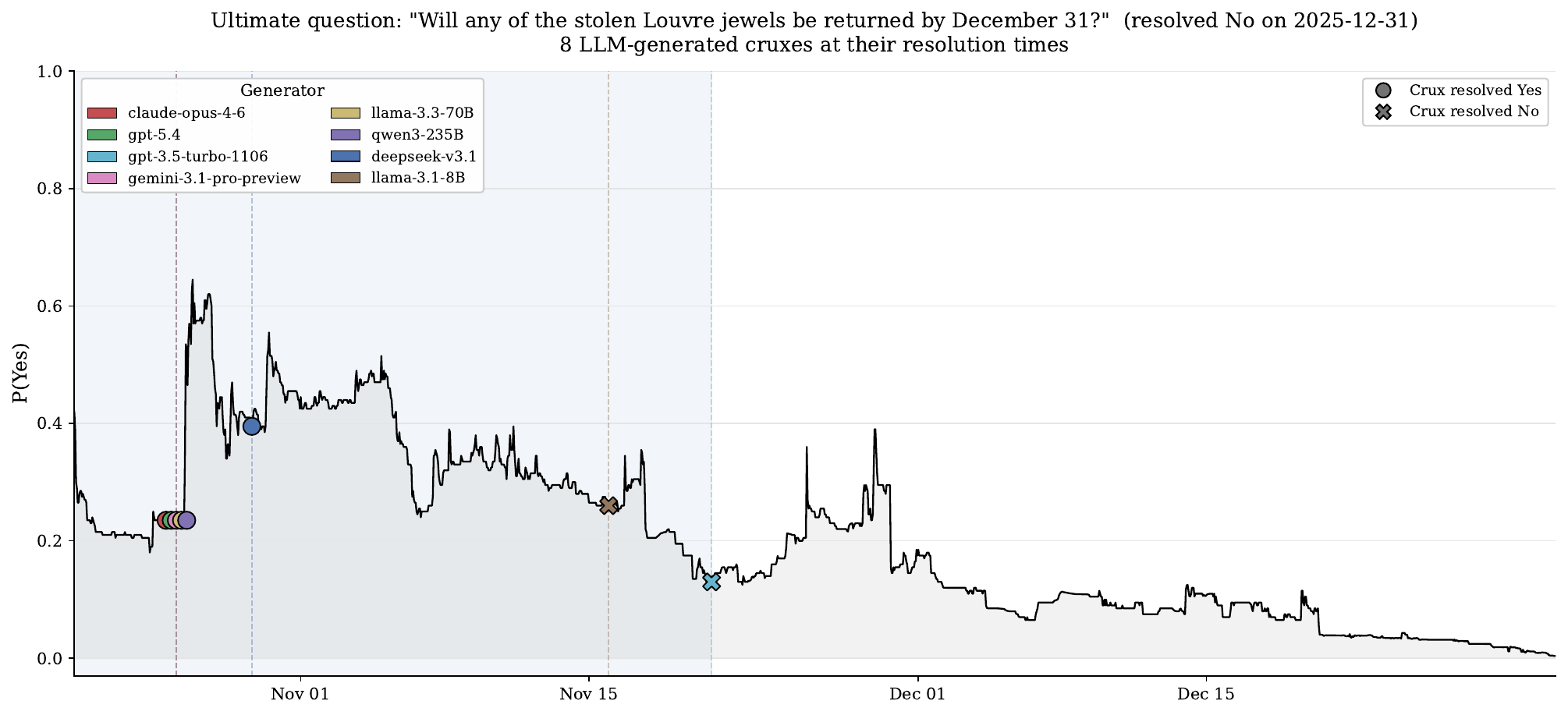}
    \caption{Price history of the ``stolen Louvre jewels'' ultimate. The shaded band is the 30-day evaluation window. Markers show each crux's resolution time, and fanned horizontally where several share a timestamp.}
    \label{fig:example_1}
\end{figure}

Figure~\ref{fig:example_1} shows that the ultimate market price rose from $p \approx 0.24$ to $p \approx 0.43$ on 2025-10-25, when French authorities announced the arrest of two suspects in the theft. From Table~\ref{tab:example_1}, five models ask for an announced arrest or charging, capturing this update and getting a high $\mathrm{KL}(\hat p, \hat p') = 0.135$. The stricter formulation from \texttt{DeepSeek-V3.1}, however, requires a suspect to be arrested \emph{and} charged. Since formal charges came four days later, by which time the market had already priced them in, this crux scores a near-zero KL. Meanwhile, \texttt{Llama-3.1-8B-Instruct} and \texttt{GPT-3.5-Turbo} ask about the jewels being recovered or appearing on the public market, neither of which happened during the window, so both resolve No at their deadlines with KL${\approx}0$.

\subsection{Example 2: NewJeans' return}
\textbf{Ultimate question:} ``Will NewJeans perform again in 2025?''

\textbf{Evaluation window:} 2025-10-31 to 2025-11-30

\begin{table*}[htbp]
\caption{Cruxes generated for ``Will NewJeans perform again in 2025?'' with their realized $\mathrm{KL}(\hat p, \hat p')$.}
\label{tab:example_2}
\centering
\small
\begin{tabular}{p{3cm} p{9.8cm} p{1.2cm}}
\toprule
\textbf{Model} & \textbf{Proposed Crux} & \textbf{KL} \\
\midrule

GPT-5.4 &
Will ADOR or NewJeans publicly announce between Oct.\ 31 and Nov.\ 30, 2025 that they have begun, resumed, or scheduled formal talks specifically aimed at restarting NewJeans group activities? &
0.5023 \\

Qwen3-235B-A22B &
Will ADOR announce new group activities for NewJeans between Oct.\ 31 and Nov.\ 30, 2025? &
0.5023 \\

Llama-3.3-70B &
Will NewJeans or their members make a public statement about their plans for 2025 between Oct.\ 31 and Nov.\ 30, 2025? &
0.5023 \\

Llama-3.1-8B-Instruct &
Will NewJeans' members contradict their recent court statement on their relationship with ADOR by engaging in any form of public activity as NewJeans, including interviews, social-media posts, or public appearances, between Oct.\ 31 and Nov.\ 20, 2025? &
0.5023 \\

Claude-Opus-4.6 &
Will NewJeans, or their representatives, file a formal appeal of the October 30, 2025 Seoul Central District Court ruling by Nov.\ 30, 2025? &
0.0539 \\

Gemini-3.1-Pro-Preview &
Will ADOR, HYBE, or NewJeans officially announce any new public group activities, such as a music release, fan event, or live performance, between Oct.\ 31 and Nov.\ 30, 2025? &
0.0007 \\

DeepSeek-V3.1 &
Will ADOR announce any NewJeans concert dates or tour plans between Oct.\ 31 and Nov.\ 30, 2025? &
0.0007 \\

GPT-3.5-Turbo &
Will NewJeans hold an official live performance as a group before Nov.\ 30, 2025? &
0.0007 \\

\bottomrule
\end{tabular}
\end{table*}

\begin{figure}[htbp]
    \centering
    \includegraphics[width=\linewidth]{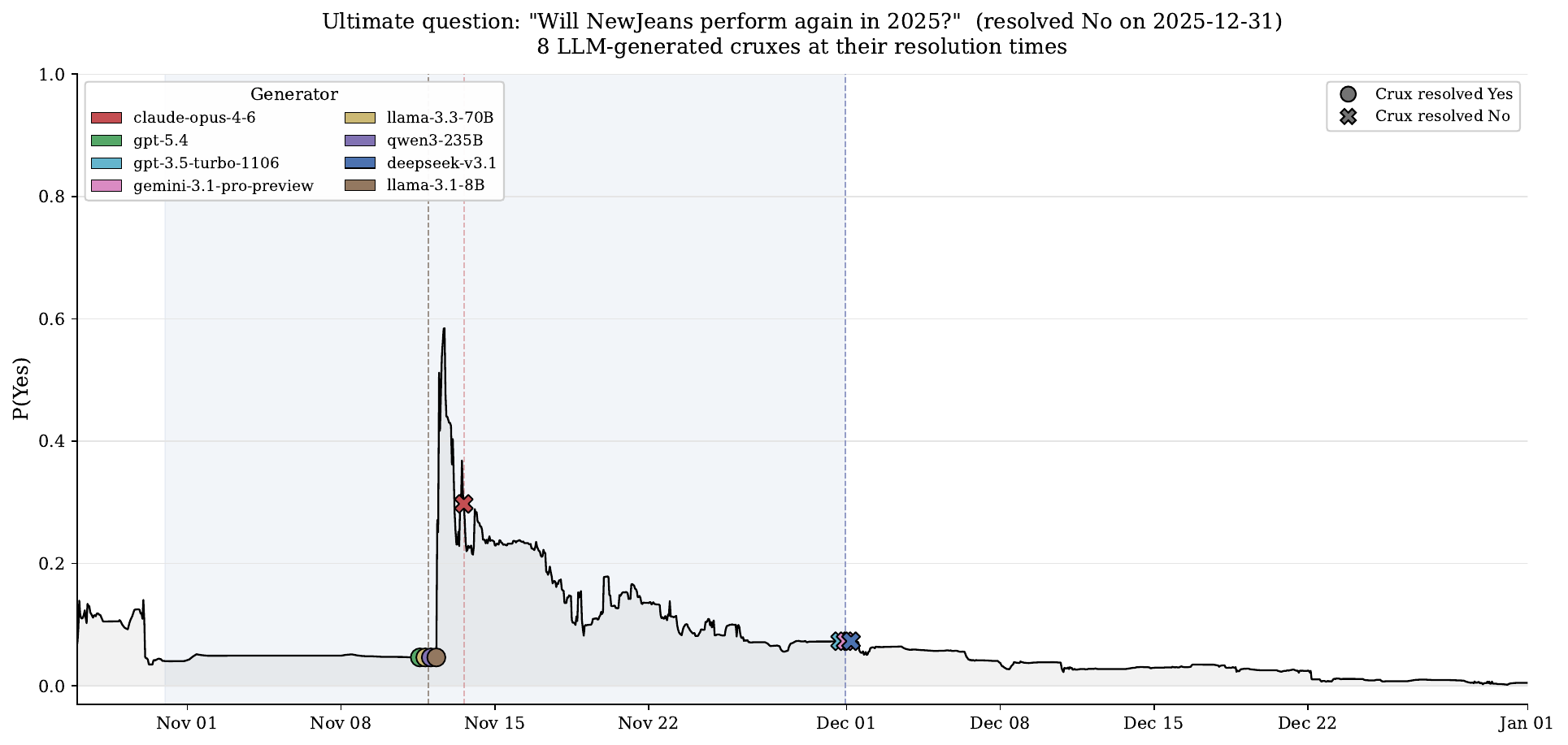}
    \caption{Price history of the ``NewJeans' return'' ultimate. The shaded band is the 30-day evaluation window. Markers show each crux's resolution time and are fanned horizontally where several share a timestamp.}
    \label{fig:example_2}
\end{figure}

Before the evaluation window, NewJeans had been engaged in a year-long contract dispute with ADOR, during which the members declared their contracts terminated and later suspended activities after a court prohibited independent work. Whether the group would signal a return to activities under ADOR is therefore highly informative for whether they would perform again in 2025. As shown in Table~\ref{tab:example_2}, \texttt{GPT-5.4}, \texttt{Qwen3-235B-A22B}, \texttt{Llama-3.3-70B}, and \texttt{Llama-3.1-8B-Instruct} all focus on whether NewJeans would resume group activities or make public statements about their future plans. The market movement supports these models' choice of crux: the November 12 return announcement coincides with a sharp increase in the probability that NewJeans would perform again, resulting in high KL scores (Figure~\ref{fig:example_2}). \texttt{Claude-Opus-4.6} instead asks whether the members will appeal the earlier court ruling, which resolves No and has a smaller market update. By contrast, \texttt{Gemini-3.1-Pro-Preview}, \texttt{DeepSeek-V3.1}, and \texttt{GPT-3.5-Turbo} require more specific downstream events such as concert plans. None occurs by their November 30 deadlines, so their cruxes resolve without a meaningful market reaction and receive KL${\approx}0$. Therefore, we see that high-KL cruxes must identify not only the right event, but also the appropriate level of specificity.

\section{Model access and decoding parameters}
\label{app:models}

Table~\ref{tab:model-access} lists each model's provider and exact model identifier. Open-weight generators (\texttt{Llama-3.3-70B}, \texttt{DeepSeek-V3.1}, \texttt{Qwen3-235B-A22B}) are accessed through the Together AI inference API\footnote{\url{https://docs.together.ai/docs/serverless-models\#chat-models}}, and \texttt{Llama-3.1-8B-Instruct} runs locally using vLLM \cite{kwon2023efficient} (on a single NVIDIA A100 80GB GPU). Across all providers we use a \texttt{max\_tokens} budget of 16{,}000. We pass each provider's default sampling parameters. Web search is disabled for all generators and forecasters. The resolver (\texttt{Claude-Sonnet-4.6}) is run with the same 16k-token budget and is given the AskNews search tool with a cap of 10 calls per crux.

\begin{table}[htbp]
\caption{Models used in our experiments: eight generators produce candidate cruxes, one resolver assigns binary outcomes and initial resolution timestamps, a verifier checks and aligns those timestamps, and three forecasters are used in the downstream forecasting experiment in Section \ref{sec:downstream-forecasting}.}
\label{tab:model-access}
\centering
\small
\setlength{\tabcolsep}{4pt}
\begin{tabular}{l l l l}
\toprule
\textbf{Role} & \textbf{Model} & \textbf{Provider} & \textbf{Identifier} \\
\midrule
Generator   & Claude-Opus-4.6 \cite{anthropic2026opus46}        & Anthropic    & \texttt{claude-opus-4-6} \\
Generator   & Gemini-3.1-Pro-Preview \cite{google2026gemini31pro} & Google       & \texttt{gemini-3.1-pro-preview} \\
Generator   & GPT-5.4  \cite{openai2026gpt54}              & OpenAI       & \texttt{gpt-5.4} \\
Generator   & DeepSeek-V3.1 \cite{deepseek2025deepseekv31release}          & Together     & \texttt{deepseek-ai/DeepSeek-V3.1} \\
Generator   & Qwen3-235B-A22B \cite{qwen3technicalreport}        & Together     & \texttt{Qwen/Qwen3-235B-A22B-Instruct-2507-tput} \\
Generator   & Llama-3.3-70B  \cite{meta2024llama33modelcard}        & Together     & \texttt{meta-llama/Llama-3.3-70B-Instruct-Turbo} \\
Generator   & Llama-3.1-8B-Instruct  \cite{grattafiori2024llama} & vLLM (local) & \texttt{meta-llama/Llama-3.1-8B-Instruct} \\
Generator   & GPT-3.5-Turbo   \cite{openai2024gpt35turbo}       & OpenAI       & \texttt{gpt-3.5-turbo-1106} \\
\midrule
Resolver    & Claude-Sonnet-4.6 \cite{anthropic2026sonnet46}      & Anthropic    & \texttt{claude-sonnet-4-6} \\
Verifier    & Claude-Sonnet-5 \cite{anthropic2026sonnet5}      & Anthropic    & \texttt{claude-sonnet-5} \\
\midrule
Forecaster  & Claude-Haiku-4.5 \cite{anthropic2025claudehaiku45}       & Anthropic    & \texttt{claude-haiku-4-5-20251001} \\
Forecaster  & Gemini-3-Flash-Preview \cite{google2025gemini3flashmodelcard} & Google       & \texttt{gemini-3-flash-preview} \\
Forecaster  & GPT-5-Mini \cite{openai2025gpt5mini} & OpenAI       & \texttt{gpt-5-mini} \\
\bottomrule
\end{tabular}
\end{table}

\section{Additional metric results}
\label{app:additional-metrics}

Table~\ref{tab:additional-metric-results} reports results under the primary KL estimator, the alternative entropy-reduction estimator, and two additional belief-update metrics. The first additional metric is squared probability change, $\Delta p^2 = (p' - p)^2$, which remains a valid VOI estimator but under the variance-based uncertainty function $f(p) = p(1-p)$ rather than entropy \citep{frankel2019quantifying}. The second additional metric is absolute probability change, $|\Delta p| = |p' - p|$. This metric is not a VOI estimator, but it provides an intuitive diagnostic of the magnitude of the observed market update, regardless of direction. Empirically, the metrics $\Delta p^2$ and $|\Delta p|$ broadly track the $\widehat{\mathrm{VOI}}^{\mathrm{KL}}$ ranking. By contrast, $\widehat{\mathrm{VOI}}^{\Delta H}$ produces a noticeably different ordering, reflecting the finite-sample behavior discussed in Section~\ref{app:estimatorchoice}.

\begin{table}[htbp]
\caption{Additional metric results by model ($n=293$). $\widehat{\mathrm{VOI}}^{\mathrm{KL}}$ and $\widehat{\mathrm{VOI}}^{\Delta H}$ are measured in bits. Bold indicates the best-performing model. Significance is relative to the random baseline: $^{*}\,p<0.05$, $^{**}\,p<0.01$.}
\label{tab:additional-metric-results}
\centering
\small
\setlength{\tabcolsep}{4pt}
\resizebox{\textwidth}{!}{%
\begin{tabular}{l r@{}l l c r@{}l l c r@{}l l c r@{}l l c}
\toprule
Model & \multicolumn{2}{c}{$\widehat{\mathrm{VOI}}^{\mathrm{KL}}$} & 95\% CI & Rank
      & \multicolumn{2}{c}{$\widehat{\mathrm{VOI}}^{\Delta H}$} & 95\% CI & Rank
      & \multicolumn{2}{c}{$|\Delta p|$} & 95\% CI & Rank
      & \multicolumn{2}{c}{$\Delta p^2$} & 95\% CI & Rank \\
\midrule

Claude-Opus-4.6
& \textbf{0.016} & $^{**}$
& [0.009,\,0.023] & 1
& \textbf{0.017} & $\phantom{^{**}}$
& [0.006,\,0.029] & 1
& \textbf{0.027} & $^{**}$
& [0.021,\,0.034] & 1
& \textbf{0.004} & $^{*}\phantom{*}$
& [0.002,\,0.006] & 1 \\

Gemini-3.1-Pro-Preview
& 0.014 & $^{*}\phantom{*}$
& [0.007,\,0.020] & 2
& 0.011 & $\phantom{^{**}}$
& [-0.001,\,0.023] & 3
& 0.024 & $^{*}\phantom{*}$
& [0.019,\,0.030] & 2
& 0.003 & $^{*}\phantom{*}$
& [0.001,\,0.005] & 2 \\

GPT-5.4
& 0.013 & $^{*}\phantom{*}$
& [0.007,\,0.019] & 3
& 0.009 & $\phantom{^{**}}$
& [-0.002,\,0.020] & 6
& 0.023 & $\phantom{^{**}}$
& [0.017,\,0.028] & 3
& 0.003 & $\phantom{^{**}}$
& [0.001,\,0.005] & 3 \\

DeepSeek-V3.1
& 0.012 & $\phantom{^{**}}$
& [0.003,\,0.020] & 4
& 0.014 & $\phantom{^{**}}$
& [0.004,\,0.025] & 2
& 0.020 & $\phantom{^{**}}$
& [0.014,\,0.026] & 5
& 0.003 & $\phantom{^{**}}$
& [0.000,\,0.005] & 4 \\

Qwen3-235B-A22B
& 0.012 & $\phantom{^{**}}$
& [0.005,\,0.018] & 5
& 0.007 & $\phantom{^{**}}$
& [-0.003,\,0.017] & 8
& 0.021 & $\phantom{^{**}}$
& [0.016,\,0.027] & 4
& 0.003 & $\phantom{^{**}}$
& [0.001,\,0.004] & 5 \\

Llama-3.3-70B
& 0.010 & $\phantom{^{**}}$
& [0.005,\,0.016] & 6
& 0.009 & $\phantom{^{**}}$
& [-0.001,\,0.019] & 7
& 0.019 & $\phantom{^{**}}$
& [0.014,\,0.024] & 7
& 0.002 & $\phantom{^{**}}$
& [0.001,\,0.004] & 6 \\

Llama-3.1-8B-Instruct
& 0.010 & $\phantom{^{**}}$
& [0.005,\,0.015] & 7
& 0.010 & $\phantom{^{**}}$
& [0.000,\,0.021] & 5
& 0.019 & $\phantom{^{**}}$
& [0.015,\,0.024] & 8
& 0.002 & $\phantom{^{**}}$
& [0.001,\,0.003] & 8 \\

GPT-3.5-Turbo
& 0.009 & $\phantom{^{**}}$
& [0.005,\,0.013] & 8
& 0.011 & $\phantom{^{**}}$
& [0.001,\,0.020] & 4
& 0.019 & $\phantom{^{**}}$
& [0.015,\,0.024] & 6
& 0.002 & $\phantom{^{**}}$
& [0.001,\,0.003] & 7 \\

\midrule

Random baseline
& 0.007 & $\phantom{^{**}}$
& [0.006,\,0.009] & ---
& 0.009 & $\phantom{^{**}}$
& [0.007,\,0.010] & ---
& 0.019 & $\phantom{^{**}}$
& [0.017,\,0.021] & ---
& 0.002 & $\phantom{^{**}}$
& [0.001,\,0.002] & --- \\

Oracle
& 0.129 & $\phantom{^{**}}$
& [0.100,\,0.157] & ---
& 0.199 & $\phantom{^{**}}$
& [0.181,\,0.217] & ---
& 0.131 & $\phantom{^{**}}$
& [0.118,\,0.145] & ---
& 0.031 & $\phantom{^{**}}$
& [0.024,\,0.038] & --- \\

\bottomrule
\end{tabular}%
}
\end{table}

For small belief updates, let $\delta = p'-p$. A Taylor expansion gives
\[
\operatorname{KL}(p,p+\delta)
=
\frac{\delta^2}{2\ln(2)p(1-p)}
+
O(\delta^3),
\]
showing that KL is locally proportional to squared price movement. Since both $|\Delta p|$ and $\Delta p^2$ are monotone in the update size $|\delta|$, they often induce similar empirical rankings despite lacking the information-theoretic grounding of KL.

\section{Averaging-window robustness for belief estimation}
\label{app:belief-window-robustness}
For each resolved crux $Q_{ij}$ with resolution time $t_{ij}$, let $P_i(t)$ denote the ``Yes'' price of the ultimate market $U_i$ at time $t$. For an averaging window of $W$ hours, we estimate the prior and posterior beliefs as:
\[
\hat p_{ij}^{(W)}
=
\frac{1}{W}\sum_{h=1}^{W} P_i(t_{ij}-h),
\qquad
\hat p_{ij}^{\prime (W)}
=
\frac{1}{W}\sum_{h=1}^{W} P_i(t_{ij}+h).
\]

\begin{figure}[htbp]
  \centering
  \includegraphics[width=0.7\linewidth]{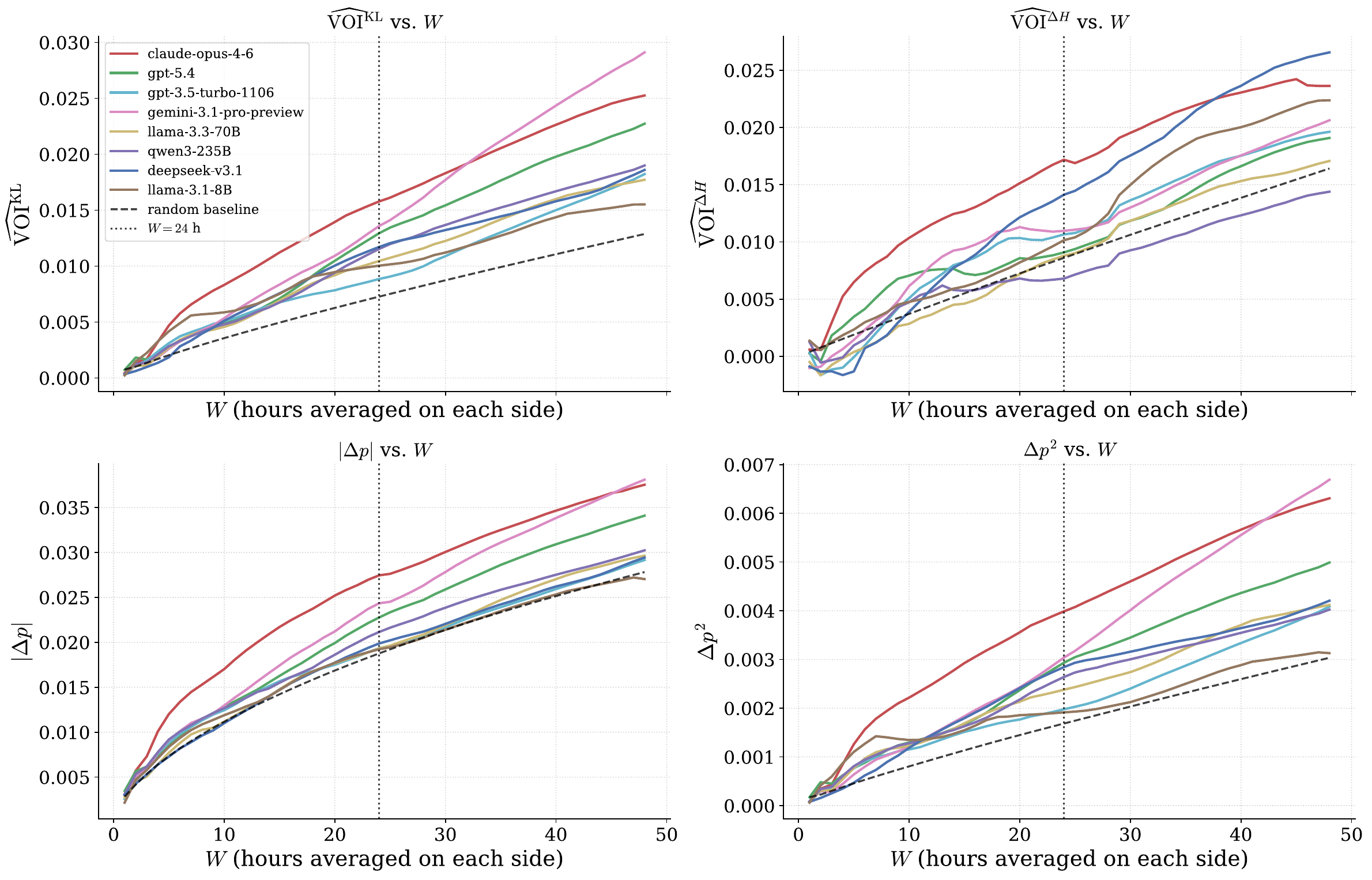}
  \caption{Different metrics vs.\ averaging window $W$. Dashed black curve is the per-hour random baseline.}
  \label{fig:window-mean}
\end{figure}

\begin{figure}[htbp]
  \centering
  \includegraphics[width=0.9\linewidth]{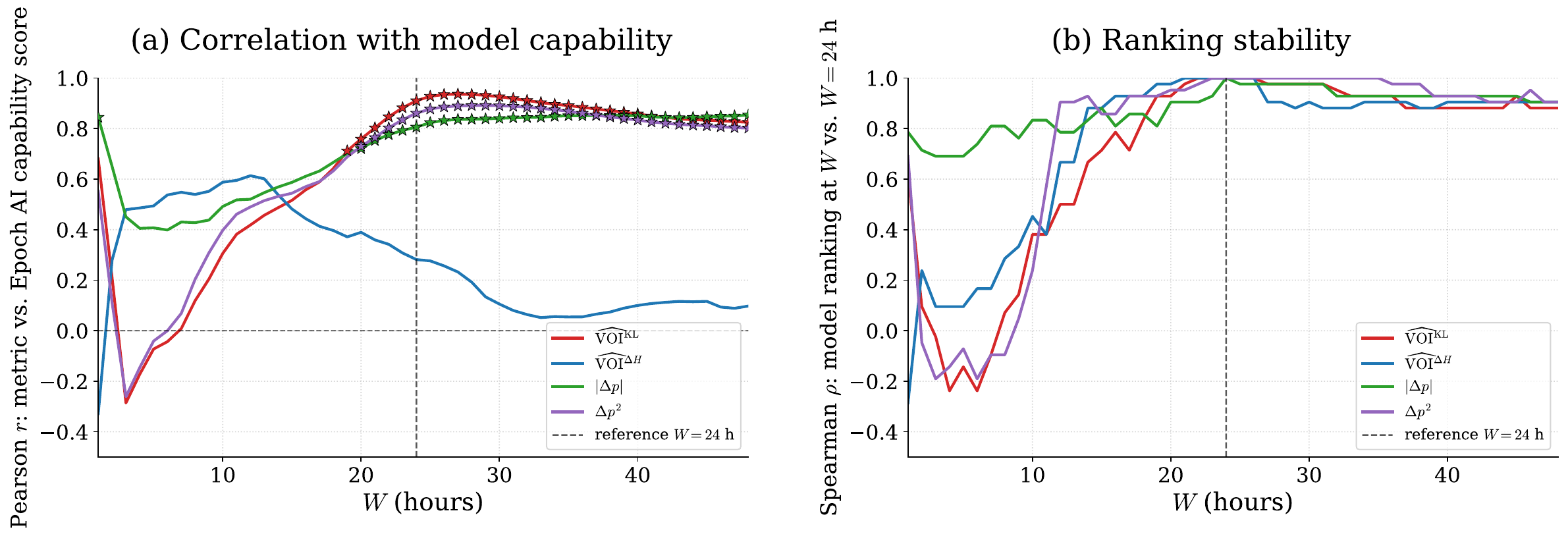}
  \caption{Window-size robustness across $W \in [1,48]$h. (a) Pearson $r$ between each metric's per-generator mean and the Epoch~AI capability score ($\star$ means $p<0.05$). (b) Spearman $\rho$ between the generator ranking at the reference $W=24$h and at every other $W$.}
  \label{fig:window-robustness}
\end{figure}

Our main results use $W=24$ hours. Sweeping $W \in [1,48]$h, Figures~\ref{fig:window-mean} and~\ref{fig:window-robustness} show that the correlation between Epoch~AI capability and each of $\widehat{\mathrm{VOI}}^{\mathrm{KL}}$, $|\Delta p|$, and $\Delta p^2$ remains relatively stable, and generator rankings are preserved except at the smaller $W$s, where short-term price noise may affect the estimation.

\section{Monte Carlo simulation details}
\label{app:monte-carlo}

\subsection{Overview}

We conduct a Monte Carlo study to compare the sampling variance of two VOI estimators: a KL-divergence and an entropy-difference. The goal is to evaluate their relative statistical efficiency under controlled belief-update regimes, where the informativeness of signals can be systematically varied.

Each simulation cell corresponds to a fixed choice of belief distributions and signal informativeness ($F_p, F_q, \alpha$), and within each cell we estimate the sampling variance of each estimator across repeated replications ($R$).

\subsection{Data generating process}

In each replication, we simulate a dataset of size $n$ consisting of ultimate questions and crux resolutions. The data are generated as $
U_i \sim \mathrm{Bern}(p), X_i \sim \mathrm{Bern}(q), Z_i \sim \mathrm{Bern}(\alpha),$ where $U_i$ is the ultimate question, $X_i$ is an independent noise draw, and $Z_i$ determines whether the crux resolution reveals truth or noise.

The crux resolution is defined as
\begin{equation}
\label{eq:mixturemodel}
Q_i =
\begin{cases}
U_i & \text{if } Z_i = 1, \\
X_i & \text{if } Z_i = 0.
\end{cases}
\end{equation}

Thus, with probability $\alpha$, the crux resolution reveals the true outcome, and with probability $1-\alpha$, it reflects unrelated noise. The parameter $\alpha \in [0,1]$ therefore controls the informativeness of the crux, ranging from pure noise ($\alpha=0$) to perfect information ($\alpha=1$).

For each replication, the parameters $p$ and $q$ are drawn from distributions $F_p$ and $F_q$, respectively.

\subsection{Posterior computation}

We compute posterior beliefs $P(U=1 \mid Q)$ in closed form via Bayes' rule. The likelihoods are:

\begin{align*}
P(Q=1 \mid U=1) &= \alpha + (1-\alpha)q, \\
P(Q=1 \mid U=0) &= (1-\alpha)q, \\
P(Q=0 \mid U=1) &= (1-\alpha)(1-q), \\
P(Q=0 \mid U=0) &= \alpha + (1-\alpha)(1-q).
\end{align*}

Applying Bayes' rule yields:
\[
P(U=1 \mid Q=1) = 
\frac{p[\alpha + (1-\alpha)q]}
{p[\alpha + (1-\alpha)q] + (1-p)(1-\alpha)q},
\]
\[
P(U=1 \mid Q=0) = 
\frac{p(1-\alpha)(1-q)}
{p(1-\alpha)(1-q) + (1-p)[\alpha + (1-\alpha)(1-q)]}.
\]

\subsection{VOI estimators}

For each observation $i$, we compute two measures of information gain:

\paragraph{KL-based estimator}
\[
IG_i^{KL} = D_{KL}\left(\mathrm{Bern}(P(U=1 \mid Q_i)) \,\|\, \mathrm{Bern}(p)\right)
\]

\paragraph{Entropy-difference estimator}
\[
IG_i^{\Delta H} = H(p) - H(P(U=1 \mid Q_i)).
\]

We aggregate within each replication:
\begin{align*}
\widehat{\mathrm{VOI}}^{KL}_r &= \frac{1}{n} \sum_{i=1}^n IG_i^{KL}, \\
\widehat{\mathrm{VOI}}^{\Delta H}_r &= \frac{1}{n} \sum_{i=1}^n IG_i^{\Delta H}.
\end{align*}

\subsection{Simulation design}

We evaluate estimator performance across a grid of simulation settings:

\begin{itemize}
    \item \textbf{Belief distributions:} 11 combinations of $(F_p, F_q)$, including uniform, symmetric Beta (e.g., $\mathrm{Beta}(2,2)$, $\mathrm{Beta}(5,5)$), skewed Beta (e.g., $\mathrm{Beta}(2,5)$, $\mathrm{Beta}(5,2)$), and Jeffreys prior $\mathrm{Beta}(0.5, 0.5)$,
    \item \textbf{Informativeness levels:} $\alpha \in \{0, 0.1, \dots, 1\}$,
    \item \textbf{Replications per cell:} $R = 1000$,
    \item \textbf{Sample size:} $n = 200$.
\end{itemize}

\begin{figure}[htbp]
    \centering
    \includegraphics[width=0.5\linewidth]{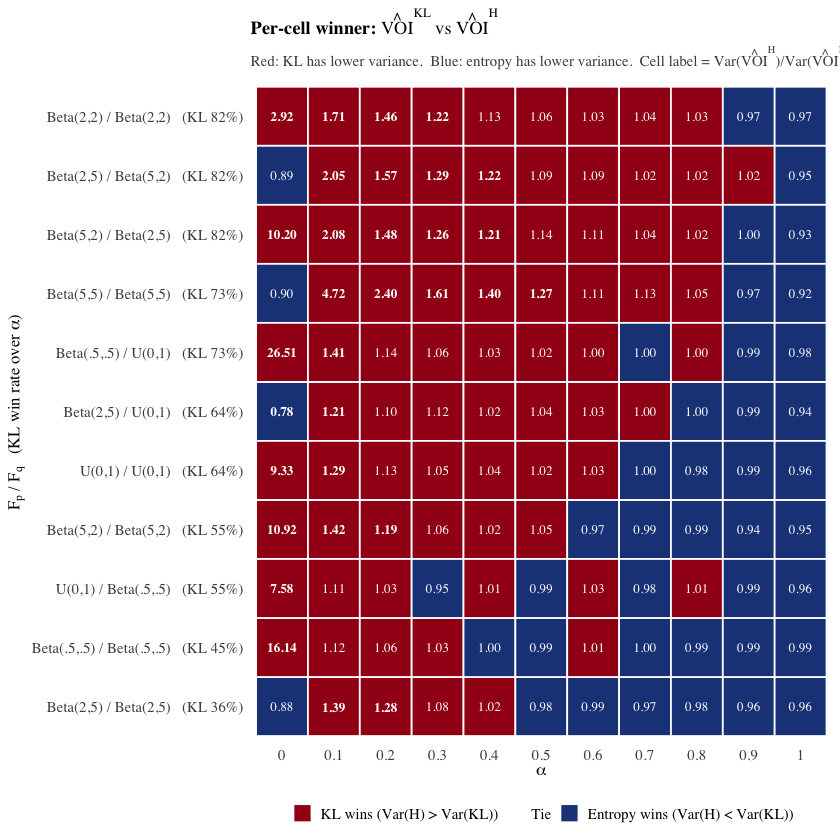}
    \caption{Where KL has lower Monte Carlo sampling variance than $\Delta H$. Heatmap over the swept $(F_p, F_q, \alpha)$ grid: \textcolor{red}{red} cells are $(F_p, F_q, \alpha)$ at which $\mathrm{Var}(\widehat{\mathrm{KL}}) < \mathrm{Var}(\widehat{\Delta H})$ (KL optimal); \textcolor{blue}{blue} otherwise. Cell winners come from an $F$-ratio test on the two replication-level variance estimates and are stable at our $R$, though variance ratios sit within a few percent of 1. The KL advantage concentrates in the low-$\alpha$ regimes ($\alpha \in [0.1, 0.6]$). The boundary cases at $\alpha \in \{0, 1\}$ are degenerate — both estimators reduce to a constant.}
    \label{fig:montecarlosim}
\end{figure}

This yields a total of 121 simulation cells and 121,000 Monte Carlo replications, as seen in Figure~\ref{fig:montecarlosim}.

\subsection{Variance estimation and comparison}

Within each simulation cell, we compute the sample variance of each estimator across replications:
\begin{align*}
\widehat{\mathrm{Var}}_{KL} &= \mathrm{Var}(\widehat{\mathrm{VOI}}^{KL}_r), \\
\widehat{\mathrm{Var}}_{\Delta H} &= \mathrm{Var}(\widehat{\mathrm{VOI}}^{\Delta H}_r).
\end{align*}

We define a \emph{win} for the KL estimator as: $
\widehat{\mathrm{Var}}_{KL} < \widehat{\mathrm{Var}}_{\Delta H},$ and the overall win rate as:
$
\text{WinRate}_{KL} = \frac{\#\{\text{cells where KL wins}\}}{\#\{\text{cells}\}}.
$ We also report the distribution of the variance ratio:
$
\frac{\widehat{\mathrm{Var}}_{\Delta H}}{\widehat{\mathrm{Var}}_{KL}},
$
including its median and geometric mean, to capture the magnitude of differences beyond binary comparisons.

\subsection{Results}

Across 121 simulation cells (Table~\ref{tab:kl-win-rate}):

\begin{itemize}
    \item KL has lower variance in 78 cells (64.5\%),
    \item Entropy has lower variance in 43 cells (35.5\%),
    \item Median variance ratio: $1.031$,
    \item Geometric mean variance ratio: $1.247$.
\end{itemize}

The advantage of KL is most pronounced in low-informativeness regimes (small $\alpha$), where posterior updates are small. Entropy-difference estimators can outperform KL in a minority of regimes, particularly under skewed parameterizations or larger updates.

\subsection{Sanity checks}

The simulation satisfies key consistency checks:

\begin{itemize}
    \item $\alpha = 0$: $Q$ is independent of $U$, implying $\mathrm{VOI} \approx 0$,
    \item $\alpha = 1$: $Q = U$, yielding maximal information,
    \item VOI increases monotonically in $\alpha$.
\end{itemize}

\subsection{Caveats}

The win-rate metric is binary and does not reflect the magnitude of variance differences. Additionally, variance estimates are subject to Monte Carlo noise, so near-ties may vary across random seeds. To address this, we report summary statistics of the variance ratio.

Finally, the boundary cases $\alpha = 0$ and $\alpha = 1$ are degenerate, with both estimators exhibiting near-zero variance.

\begin{table}[htbp]
\centering
\caption{Per-combination win rate for KL. Fraction of $\alpha$-values (per row $n_\alpha = 11$) at which $\mathrm{Var}(\widehat{\mathrm{KL}}) < \mathrm{Var}(\widehat{\Delta H})$, with median variance ratio $\mathrm{Var}(\widehat{\Delta H})/\mathrm{Var}(\widehat{\mathrm{KL}})$. Under a two-sided sign test of $H_0: P(\mathrm{KL\ wins}) = 0.5$, the strongest rows ($81.8\%$) yield $p \approx 0.065$. Variance ratios cluster within $[0.98, 1.14]$, so per-combination effects are small. Across all 11 combinations, $9$ favor KL (sign test, $p \approx 0.065$), indicating a directional advantage.}
\label{tab:kl-win-rate}
\begin{tabular}{lccc}
\toprule
\textbf{Combination ($F_p / F_q$)} & \textbf{$n_\alpha$} & \textbf{KL win rate (\%)} & \textbf{Median Var(H)/Var(KL)} \\
\midrule
beta(2,2) / beta(2,2)   & 11 & 81.8 & 1.06 \\
beta(2,5) / beta(5,2)   & 11 & 81.8 & 1.09 \\
beta(5,2) / beta(2,5)   & 11 & 81.8 & 1.14 \\
beta(5,5) / beta(5,5)   & 11 & 72.7 & 1.13 \\
jeffreys / uniform      & 11 & 72.7 & 1.02 \\
beta(2,5) / uniform     & 11 & 63.6 & 1.02 \\
uniform / uniform       & 11 & 63.6 & 1.03 \\
beta(5,2) / beta(5,2)   & 11 & 54.5 & 1.02 \\
uniform / jeffreys      & 11 & 54.5 & 1.01 \\
jeffreys / jeffreys     & 11 & 45.5 & 0.997 \\
beta(2,5) / beta(2,5)   & 11 & 36.4 & 0.980 \\
\bottomrule
\end{tabular}
\end{table}

\section{Empirical calibration of signal informativeness}
\label{app:alpha-estimation}

In addition to the Monte Carlo analysis, we estimate the level of signal informativeness $\alpha$ implied by observed data. This provides an empirical calibration of the regimes studied in the simulation and helps assess whether the parameter values considered are realistic.

\subsection{Identification of $\alpha$}

In the mixture model (as above in equation (\ref{eq:mixturemodel})), the conditional distribution of the crux resolution $Q$ given the ultimate question $U$ satisfies:
\begin{align*}
P(Q=1 \mid U=1) &= \alpha + (1-\alpha)q, \\
P(Q=1 \mid U=0) &= (1-\alpha)q.
\end{align*}

Therefore,
\[
\alpha = P(Q=1 \mid U=1) - P(Q=1 \mid U=0) = \mathrm{TPR} - \mathrm{FPR}.
\]

Thus, $\alpha$ is identified as the difference between the true positive rate and false positive rate (Youden's J statistic \cite{youden1950index}). Importantly, the dependence on $q$ cancels, so $\alpha$ can be estimated directly from empirical crux resolutions $Q$ and realized outcomes $U$ without specifying the noise distribution.

\subsection{Empirical estimation}

We estimate the informativeness parameter $\alpha$ using observed crux resolutions. In the data, each market provides:
\begin{itemize}
    \item an ultimate question $U \in \{0,1\}$ (market resolution), and
    \item a corresponding crux resolution $Q \in \{0,1\}$ (crux resolution).
\end{itemize}

This directly matches the mixture model, where $Q$ is a binary signal about $U$. We therefore estimate:
\begin{align*}
\widehat{\mathrm{TPR}} &= \mathbb{P}(Q=1 \mid U=1) = \frac{\#\{Q=1,\, U=1\}}{\#\{U=1\}}, \\
\widehat{\mathrm{FPR}} &= \mathbb{P}(Q=1 \mid U=0) = \frac{\#\{Q=1,\, U=0\}}{\#\{U=0\}}, \\
\widehat{\alpha} &= \widehat{\mathrm{TPR}} - \widehat{\mathrm{FPR}}.
\end{align*}

This estimator corresponds to Youden's $J$ statistic and provides a direct measure of how informative the crux is about the ultimate question, and using our empirical data we get Table~\ref{tab:empestofalpha}.

\subsection{Results}

Pooling the 2$\times$2 contingency of $(U, Q)$ across all eight forecasting models ($N = 2{,}152$ resolvable cruxes) yields
$$\widehat{\mathrm{TPR}} = 0.324, \qquad \widehat{\mathrm{FPR}} = 0.295, \qquad \hat{\alpha} = 0.030.$$
The standard error from a two-sample test of proportions,
$\widehat{\mathrm{SE}}(\hat\alpha) = \sqrt{\widehat{\mathrm{TPR}}(1-\widehat{\mathrm{TPR}})/n_+ + \widehat{\mathrm{FPR}}(1-\widehat{\mathrm{FPR}})/n_-} = 0.022$
($n_+ = 604$, $n_- = 1{,}548$), gives a 95\% Wald CI of $[-0.014, 0.074]$ and $p \approx 0.18$. The pooled estimate is positive but indistinguishable from zero at the 5\% level. Per-model estimates (Table~\ref{tab:empestofalpha}) span $\hat\alpha \in [-0.060, 0.123]$: only the three strongest models (\texttt{Gemini-3.1-Pro-Preview}, \texttt{GPT-5.4}, \texttt{Claude-Opus-4.6}) yield $\hat\alpha > 0.11$, and even these do not individually clear zero.

This empirical calibration places the realistic operating regime of crux-style decomposition firmly in the low-$\alpha$ band of the simulation in Appendix \ref{app:monte-carlo}, supporting the paper's focus on low-information settings when comparing VOI estimators.

\begin{table}[htbp]
\centering
\caption{Empirical estimate of crux informativeness, by model. $\hat{\alpha} = \mathrm{TPR}-\mathrm{FPR}$ (Youden's $J$); $n_+, n_-$ are the per-row counts of $U=1$ and $U=0$.}
\label{tab:empestofalpha}
\begin{tabular}{lrrrrr}
\toprule
Model & $\hat{\alpha}$ & TPR & FPR & $n_+$ & $n_-$ \\
\midrule 
Gemini-3.1-Pro-Preview         & $+0.123$ & $0.405$ & $0.282$ & $79$ & $195$ \\
GPT-5.4                        & $+0.118$ & $0.444$ & $0.327$ & $72$ & $196$ \\
Claude-Opus-4.6                & $+0.113$ & $0.436$ & $0.323$ & $78$ & $201$ \\
GPT-3.5-Turbo                  & $+0.042$ & $0.222$ & $0.180$ & $81$ & $194$ \\
DeepSeek-V3.1                  & $-0.014$ & $0.181$ & $0.195$ & $72$ & $190$ \\
Llama-3.3-70B   & $-0.034$ & $0.380$ & $0.414$ & $71$ & $181$ \\
Llama-3.1-8B-Instruct          & $-0.050$ & $0.260$ & $0.309$ & $77$ & $194$ \\
Qwen3-235B-A22B    & $-0.060$ & $0.270$ & $0.330$ & $74$ & $197$ \\
\midrule
\textbf{Pooled (micro)}        & $\mathbf{+0.030}$ & $\mathbf{0.324}$ & $\mathbf{0.295}$ & $\mathbf{604}$ & $\mathbf{1{,}548}$ \\
\bottomrule
\end{tabular}
\end{table}

\section{Cosine similarity and operationalization differences}
\label{app:cosine}

Figure~\ref{fig:sub_q_similarity_op} compares crux pairs along two operational dimensions: numeric \emph{threshold} and \emph{source kind} (structured data providers versus news outlets). It reports the within-pair match rate for each dimension by question cosine similarity. Threshold agreement remains low: even among near-paraphrases ($\cos > 0.8$), only 16\% of pairs use the same numeric cutoff. Source-kind agreement increases with semantic similarity, from 51\% to 68\%, but remains far from complete. Thus, even semantically similar cruxes often differ in both their exact criterion and the type of evidence used to resolve it.

\subsection{Source Kind Taxonomy}
\label{app:source-taxonomy}

\begin{figure}[htbp]
    \centering
    \includegraphics[width=0.9\linewidth]{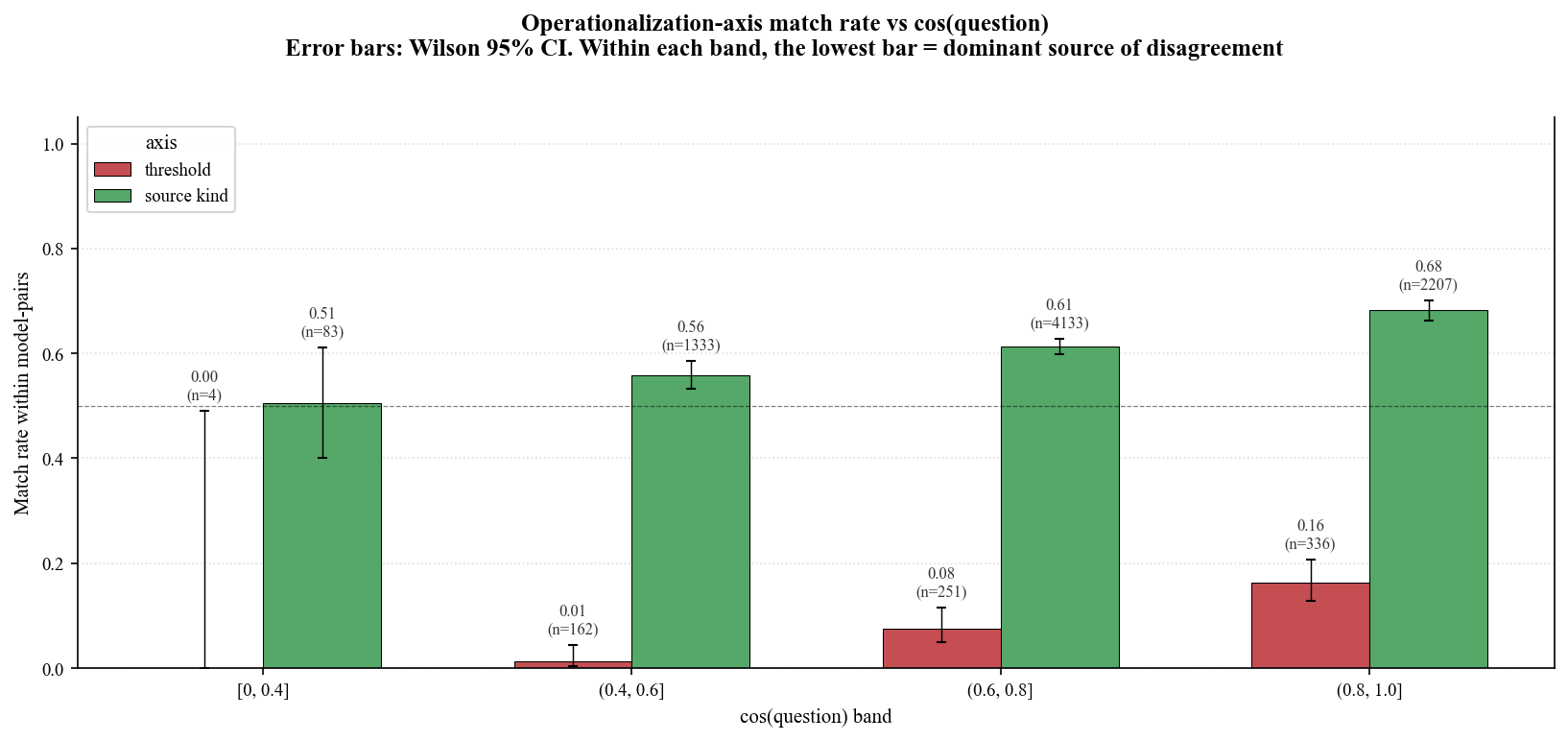}
    \caption{Operationalization-axis match rate by cosine similarity. For each question pair we record whether the two cruxes share the same numeric \emph{threshold} (when both emit one) and the same \emph{source kind} (Appendix~\ref{app:source-taxonomy}). Error bars are Wilson 95\% CIs on each binomial proportion. Even at $\cos > 0.8$, threshold agreement is only $0.16$ and source-kind agreement only $0.68$.}
    \label{fig:sub_q_similarity_op}
\end{figure}

For the operationalization analysis in Figure~\ref{fig:sub_q_similarity_op}, we extract named sources from each crux's resolution criteria via case-insensitive substring match against a fixed keyword list. Each named source is then mapped to one of two broad categories: structured and news. The \emph{source kind\_match} axis asks whether two cruxes share the same category, while \emph{source} asks semantic similarity. 

\paragraph{Structured (quantitative / official data providers).}
\begin{itemize}
    \item \textbf{Reviews and entertainment data:} Metacritic, Rotten Tomatoes, IMDb, Box Office Mojo, The-Numbers, Billboard, Spotify, Steam (incl.\ SteamDB, SteamCharts), Nielsen.
    \item \textbf{Macroeconomic data:} FRED (St.\ Louis Fed), Federal Reserve / FOMC, Bureau of Labor Statistics, Bureau of Economic Analysis, U.S.\ Treasury.
    \item \textbf{Markets and exchanges:} CME Group (incl.\ FedWatch), NYSE, NASDAQ, SEC (incl.\ EDGAR).
    \item \textbf{Crypto and prediction markets:} CoinGecko, CoinMarketCap, Polymarket, Kalshi, PredictIt.
    \item \textbf{Government and public health:} NOAA (incl.\ weather.gov, NCEI), FBI, CDC, WHO.
    \item \textbf{Reference:} Wikipedia.
\end{itemize}

\paragraph{News (journalistic outlets).}
New York Times, Wall Street Journal, Reuters, Bloomberg, Financial Times,
Associated Press, CNN, BBC, The Guardian, Washington Post, CNBC, Forbes,
Fortune, Axios, Politico, Business Insider, Yahoo Finance.

A crux mentioning at least one structured source is labeled \texttt{structured}; one mentioning only news outlets is labeled \texttt{news}; and one mentioning neither is labeled \texttt{none}. When a crux mentions both, it is labeled \texttt{structured}, since structured providers are the more committal anchor for resolution.

\section{Unresolvable questions and their classification}
\label{sec:unresov}  
We define \emph{resolvability} as whether the resolver can evaluate a crux as true or false at a specific time based on available evidence. We sample 50 unresolvable questions,\footnote{The 50 sampled questions are drawn to roughly match each model’s share of unresolvable questions in the full dataset.} and identify three main failure modes:

\begin{enumerate}[leftmargin=*]
\item \textbf{Data availability failures.} ($50.0\%)$\footnote{Percentage of the 50 sampled questions in this category. Table~\ref{tab} provides the per-model breakdown.}
The crux relies on data that is unstable, inaccessible, or unavailable at the required level of granularity, such as niche dashboards that are not retrievable, data not published at the required level of breakdown, or source-platform ambiguity. \\
\textbf{Example:} ``Will Phantom's revenue exceed \$200k on ...'' There are live dashboards like Flowscan, but news sources do not archive the fine-grained daily values required to resolve the question later.

\item \textbf{Temporal misalignment.} ($18.0\%$)
The crux is tied to an event or data release that does not occur within the specified resolution window, either because the data release/a fixed event is scheduled, or delayed to after the window closes. \\
\textbf{Example:} Delayed government reports or sports matches. Underlying event occurs within window but official data released after.

\item \textbf{Observability failures.} ($32.0\%$)
The crux depends on events that are not reliably observable, typically because they involve unverifiable negatives, exhaustive enumeration, or events whose non-occurrence is not publicly recorded. \\
\textbf{Example:} ``Will at least two women appear on TIME covers between ...'' Yes resolves from observing two covers, while No requires exhaustively checking every cover in the window.
\end{enumerate}

We note that the majority (50\%) of failures stem from data availability, which is expected as the resolver pipeline relies on news: if aggregate information is unavailable or the metric is niche, the resolver would fail.

\label{app:unresov}
\begin{table}[htbp]
\centering
\caption{Unresolvable cruxes by failure mode and model ($n = 50$). Counts of cruxes each model marked unresolvable, classified into three failure modes: \emph{data availability}, \emph{temporal}, and \emph{observability}. Data availability dominates overall ($25/50 = 50\%$), but the distribution is model-dependent: \texttt{Llama-3.3-70B} and \texttt{Qwen3-235B-A22B} most often fail due to data availability, while \texttt{Llama-3.1-8B-Instruct}'s failures are predominantly temporal. \texttt{Gemini-3.1-Pro-Preview} and \texttt{Claude-Opus-4.6} show a larger share of observability failures.}
\label{tab}
\begin{tabular}{lrrrr}
\toprule
\textbf{Model} & \textbf{Data avail.} & \textbf{Temporal} & \textbf{Observability} & \textbf{Total} \\
\midrule 
Claude-Opus-4.6 & 1 & 0 & 2 & 3 \\
Gemini-3.1-Pro-Preview & 2 & 0 & 3 & 5 \\
GPT-5.4 & 3 & 2 & 1 & 6 \\
DeepSeek-V3.1 & 4 & 1 & 3 & 8 \\
Qwen3-235B-A22B & 5 & 0 & 1 & 6 \\
Llama-3.3-70B & 8 & 0 & 3 & 11 \\
Llama-3.1-8B-Instruct & 0 & 4 & 2 & 6 \\
GPT-3.5-Turbo & 2 & 2 & 1 & 5 \\
\midrule
All & 25 & 9 & 16 & 50 \\ 
\bottomrule
\end{tabular}
\end{table}

\section{Prompts}
\subsection{Crux generation prompt}
\label{app:crux-generation-prompt}
\begin{promptbox}{Crux Generation System Prompt}
You are a helpful assistant to forecasting research scientists. The goal is to generate subquestions / cruxes for a given ultimate question.

TASK BACKGROUND:
In forecasting, we often care about answering questions that are either too complex, too broad or too far in the future to answer directly. If the outcome of a question will only be known in 20 years, we would like to have indicators for which track the world is on, with respect to the outcome of the question. For example, if the ultimate question regards nuclear safety in 2050, we would like to have indicators in the near term which could be used as indicators for the ultimate question - such as the number of nuclear power plants in the world, the number of nuclear weapons in the world, the number of nuclear accidents in the world, etc.

These indicators are called subquestions / cruxes. They are indicators that can be used to track the world's progress towards the ultimate question. We want the subquestions to be grounded in current world events that pertain to the ultimate question. Ideal subquestions are such that they are cruxes for ultimate question disagreements; i.e if 2 forecasters disagree on the outcome of the ultimate question, they are more likely to agree on the outcome of the final question once the subquestion is resolved.

TASK:
- Generate subquestions that are meaningful early indicators of the ultimate question's outcome -- not restatements of the ultimate question with an earlier end date.

GUIDELINES:

Questions:
- The question should be binary. This means that the outcome of the question should be either yes or no. The question should be formulated in such a way that forecasters answer with a probability. Hence, a question such as "will X increase or decrease" is not a good question. A better question is "Will X increase by at least 10 units before [specific date]?". A higher probability answer implies more likely 'yes', a lower probability implies more likely 'no'.
- Informative: Is this crux / subquestion likely to provide evidence for the ultimate outcome question that the user cares about? The question should have a high value of information - meaning that knowing the outcome of this question would update forecasters on their forecast of the ultimate outcome.
- Good questions usually do not have an expected forecast very near 0 or 1, because then the forecasts will not be informative. This is a general guideline, not a rule: there may be exceptions to this.
- Consider potential loopholes in the resolution that make the forecasts uninformative. I.e. if we are curious about the difference between 2 competing LLMs on a benchmark, and the benchmark becomes saturated, the question will be resolvable, but not informative (both competitors will have the same score of 100%).
- The question should be a clear and concise question that is easy to understand.
- The question should be grounded in current world events. For whatever question you want to ask, ensure you know the current state of that area to choose appropriate thresholds for the question.
- All terms in the question and resolution criteria must be clearly defined, eliminating ambiguity.
- Time-bounded: The subquestion end_date MUST be before the end date of the ultimate question. The subquestion end_date must also be AFTER the start date of the ultimate question (i.e. today's date). Both the ultimate question end date and start date will be provided to you.

Background information:
- The background information should provide enough context to fully understand the question and why it is relevant.
- The background information should include the current state of the area, including relevant thresholds for the question. For example if you are forecasting the number of nuclear weapons in the world, you should include the current number of nuclear weapons in the world. If you are forecasting the number of AI jobs, you should include the current number of AI jobs etc. These should influence the thresholds for the question.
- The background information must provide citations.
- Citations MUST include actual, clickable URLs (http/https). Do NOT output placeholder citation tokens like "citeturn5search5...". Inline the source URL next to claims or include them in a short Sources list.
- At the end of Background information, add a short "Sources:" list with each item on its own line containing the source title (or site) and full URL.

Resolution criteria:
- Aim for tight resolution criteria. The resolution criteria should leave little room for discretion in deciding the resolution. As best you can, try to limit the scope for ex-post quarrels about what really happened, and who was right.
- Define your terms. Questions of the sort "will X occur?" often hinge on how X is defined. It is therefore important to spell out your definitions with extra care. Don't worry about being too pedantic here!
- Be concrete. Try to specify precisely and in detail which steps should or shouldn't be followed when resolving the question. Examples are helpful for making these instructions concrete.
- Use authoritative sources, when possible. Good options are numerical data regularly published by a reliable publicly available source. Note that you should be sure that the sources will be available at the time of resolution, or otherwise you might want to specify alternative sources of information.
- Consider and account for edge-cases. Try to imagine scenarios for which the resolution conditions fail to cleanly apply, or cases that are just on the edge of counting towards resolution. If such scenarios or edge-cases are plausible, you should clarify how the question should resolve when such events rear their head.
- Consider fall-back criteria. When you have a resolution that should be easy to check assuming all goes well, try to handle also the case where all doesn't go well. What if the data source you specified stops being published? Is there anything else odd that might happen to make the outcome unclear?
- Consider delayed events: if the event is postponed beyond the resolution window, the question should be considered unresolvable -- specify this explicitly in the resolution criteria.
- Try to account for unknown unknowns. Think about how the resolution criteria behave when something you don't expect happens anyway.
- Well-researched: The question should be well-researched and grounded in current world events. It must provide high quality sources for the resolution criteria. The background information must provide enough context to fully understand the question and why it is relevant. Both the resolution criteria and the background information must provide citations.
- For the Resolution criteria, include the specific data source(s) you propose to use and provide the full URLs. Avoid any placeholder tokens; always include explicit links.
- Time-bounded: The question should specify when it will be resolved. MUST explicitly state the start date and end date of the subquestion, clearly defining the resolution window. For example: "This question resolves Yes if [condition] occurs between [start_date] and [end_date] (inclusive)." This ensures resolvers know the exact time window during which qualifying events must occur.
\end{promptbox}

\begin{promptbox}{Crux Generation User Prompt}
Ultimate question: {ultimate_question}
Description: {description}
Today's date: {start_date}
Ultimate question end date: {end_date}

Generate {n} subquestion(s) whose start_date is today ({start_date}) and end_date is no later than {period_end}.
IMPORTANT:
- Every subquestion end_date MUST be between {start_date} and {period_end}.
- The resolution_criteria for each subquestion MUST explicitly mention the subquestion's start_date ({start_date}) and end_date, clearly stating the window during which qualifying events count toward resolution. For example: "This question resolves Yes if [condition] occurs between {start_date} and [end_date]."
- DO NOT generate a subquestion that is just the ultimate question with an earlier deadline or tighter resolution criteria. A good subquestion asks about a different observable event that could be an indicator of the ultimate outcome.

{articles_block}
RETURN FORMAT:
Return a JSON array of dictionaries. Each dictionary MUST contain:
- question: the subquestion / crux
- background_information: background information on the question. Include citations with explicit, clickable URLs and add a short Sources list at the end.
- resolution_criteria: the resolution criteria. Must explicitly state the start_date and end_date defining the resolution window. You must provide links (full URLs) to the sources which you propose to use for the resolution criteria. Do not include placeholder citation tokens.
- start_date: "{start_date}" (today's date, MUST be exactly this value)
- end_date: the date when the question will be resolved (in UTC). MUST be between {start_date} and {period_end}.
- rationale: a rationale for why this question is relevant to the ultimate question

Return ONLY the JSON array, no other text.
\end{promptbox}

\subsection{Crux resolution prompts}
\label{app:crux-resolution-prompt}

\begin{promptbox}{Initial Resolution System Prompt}
You are a research assistant that determines whether forecasting questions have resolved Yes or No based on news articles. Use the asknews_search tool to find relevant news articles and determine the resolution.
\end{promptbox}

\begin{promptbox}{Initial Resolution User Prompt}
Determine whether the following binary forecasting question has resolved Yes or No.

Question: {question}
Resolution criteria: {resolution_criteria}
Start date: {start_date}
End date: {end_date}

You are a careful, high-precision resolution assistant. Your goal is to determine whether the question resolved Yes or No using explicit, verifiable evidence.

IMPORTANT -- resolution rules:
- Only events that occurred BETWEEN {start_date}T00:00:00Z and {end_date}T23:59:59Z (inclusive) count.
- Prefer precision over recall. If uncertain, return resolvable=false.
- Do NOT guess, infer, or assume. Only use explicit evidence.

resolvable=false if ANY of the following:
- The underlying event or data is missing, unavailable, or not reported
- The resolution criteria are ambiguous, contradictory, or unclear
- The evidence is indirect, inferred, or does not exactly match the criteria
- A reliable first-occurrence timestamp cannot be established
- Conflicting evidence cannot be resolved
- Any reason prevents a definitive Yes/No answer

resolution=Yes:
- The resolving event explicitly occurred within the window
- If multiple qualifying events exist, use the FIRST occurrence
- You MUST verify that no earlier qualifying occurrence exists in the window

resolution=No:
- If the condition became definitively impossible before end_date -> resolution_datetime = the exact datetime it became impossible. Always use the earliest date at which a definitive No conclusion is supported by explicit evidence. "Definitively impossible" means there is explicit, public evidence of a discrete event that forecloses the Yes outcome (e.g., a deadline passed, a vote failed, a candidate withdrew) -- not merely that the probability appeared low.
- If the window closed with no earlier definitive No moment -> resolution_datetime = {end_date}

EVIDENCE REQUIREMENTS (ALL must be satisfied for resolvable=true):
1. Evidence directly matches the resolution criteria (no interpretation required)
2. Event is explicitly stated (not inferred)
3. Event is clearly within the window
4. A reliable timestamp exists (not approximate)
5. Source is high-quality OR multiple sources agree

If any condition fails -> resolvable=false

STRICT RULES FOR resolution_datetime:
- Must be an exact ISO 8601 UTC datetime
- No vague language ("around", "approximately", etc.)
- If timestamp is not explicitly available or cannot be reliably determined -> resolvable=false

Follow this procedure:

STEP 1 -- DECOMPOSE
- Identify the exact condition required for Yes
- Identify what observable evidence would confirm it
- Identify key entities, thresholds, and timing constraints

STEP 2 -- TARGETED SEARCH
- Generate multiple precise search queries including:
  * entity + metric + date
  * variations to find earliest occurrence
  * queries to disprove the event

STEP 3 -- EVENT DETECTION
- Identify candidate events
- Reject any that:
  * require inference
  * are indirect or approximate
  * do not clearly satisfy criteria

STEP 4 -- CRITERIA MATCHING
- Confirm the event EXACTLY satisfies the resolution criteria
- Reject partial or approximate matches

STEP 5 -- TIMESTAMP EXTRACTION
- Extract earliest exact timestamp of occurrence
- Reject if timestamp is vague or inferred

STEP 6 -- FIRST OCCURRENCE CHECK
- Search specifically for earlier qualifying events
- If you cannot confirm this is the first -> resolvable=false

STEP 7 -- VERIFICATION (CRITICAL)
- Attempt to disprove your conclusion:
  * look for contradictory evidence
  * check similar events outside window
  * confirm correct interpretation
- If contradiction exists -> resolvable=false

STEP 8 -- FINAL DECISION
- Yes: event occurred within window (first occurrence)
- No: condition never met within window
- Otherwise: resolvable=false

Double check before answering:
1. Evidence is direct, explicit, and within the window
2. No inference or assumptions were used
3. Timestamp is exact and reliable
4. FIRST occurrence condition is satisfied
5. No contradictory evidence exists
6. If any doubt remains -> resolvable=false

Return STRICT JSON:
{{   
    "evidence": {{
        "step1_decompose": "<exact condition for Yes, observable evidence required, key entities/thresholds/timing>",
        "step2_search_queries": "<search queries used and what each was intended to find>",
        "step3_event_detection": "<candidate events identified and any rejected with reasons>",
        "step4_criteria_matching": "<confirmation or rejection of exact criteria match -- include inline citations [source](url)>",
        "step5_timestamp_extraction": "<exact timestamp found, or why it could not be extracted>",
        "step6_first_occurrence_check": <whether an earlier qualifying occurrence was found>,
        "step7_verification": "<contradictory evidence checked, alternative interpretations considered>",
        "step8_final_decision": "<final reasoning tying all steps to the resolution>"
    }},
    "resolvable": true or false,
    "resolution": "Yes" or "No" or "",
    "resolution_datetime": "<exact UTC datetime in ISO 8601, or empty if resolvable=false>",
    "source_urls": ["url1", "url2", ...]
}}
\end{promptbox}

\begin{promptbox}{Timestamp Verification Prompt}
You are a resolution-time adjudicator for a forecasting benchmark. Several language models each proposed one crux (subquestion) for the same ultimate market question, and a resolver assigned each crux a Yes/No resolution and a timestamp. Resolver timestamps are noisy: the same real-world event sometimes received different timestamps for different models' cruxes, and some 'No' resolutions were stamped at an evidence time even though the answer was not yet determined.

Your job is to adjudicate TIMESTAMPS ONLY. Never change a Yes/No label.

RULES
1. Event grouping: identify which cruxes are resolved by the same underlying real-world event (across models). All cruxes in one event group must receive the IDENTICAL timestamp: the earliest time the resolving evidence plausibly became public, based on the evidence provided (event dates, article dates, official announcement times). Cruxes resolved by genuinely different events (e.g., an arrest vs. a later formal charging, when the crux wording requires the latter) belong to different groups even if close in time.
2. Yes resolutions: timestamp = the canonical public time of the resolving event (per its event group).
3. No resolutions: timestamp = the crux's OWN deadline, at end_date T23:59:59Z (rule "no_deadline") -- UNLESS an event strictly before the deadline made the No logically or institutionally irreversible, e.g. the subject was eliminated from the process, an authority issued a binding and final negative decision, or the only possible path to Yes was closed (rule "no_irreversible", timestamp = that event's canonical time). Announcements of delay, postponement, low likelihood, or intent are NOT irreversible: the answer remains open until the deadline. Worked example: a crux asks "will NASA complete a rehearsal before Feb 16?"; on Feb 3 the first attempt is scrubbed and NASA announces a second rehearsal is needed and the launch is moving to March. This is rule "no_deadline" (stamp Feb 16): schedules can change again, so completion before Feb 16 remained possible -- a delay announcement is evidence of No, not determination of No. Contrast: if that crux's deadline were Feb 5 and completing another attempt within 2 days were physically/procedurally impossible, rule "no_irreversible" (stamp the scrub announcement time) would apply. Apply "no_irreversible" only when you can state WHAT made Yes impossible before the deadline, not merely unlikely. Two hard requirements for "no_irreversible":
   (a) NO HINDSIGHT: you must justify impossibility using only information available AT the candidate event's time. That the Yes-event in fact never happened before the deadline (known from later evidence) is NOT an argument -- every No crux trivially satisfies that. If your justification cites what happened afterwards, it is invalid; use "no_deadline".
   (b) BINDING CONSTRAINT: the impossibility must follow from a binding physical, legal, or procedural constraint with fixed dates (e.g., candidate eliminated from a completed vote; the only scheduled session already past; a law's effective date). Repair timelines, agency scheduling projections, "requires X first" chains, and stated targets are expectations, not constraints -- they can be revised, so they are NOT irreversible.
4. If the resolver's existing timestamp already satisfies these rules, keep it.
5. Use UTC ISO format like 2026-02-03T17:00:00Z. Timestamps must lie within the crux's [start_date, end_date 23:59:59] window.
6. TIMEZONE DISCIPLINE AND TIME PRECISION: source times are often stated in local timezones (ET, PT, CET, ...) or as bare dates. Assign the event time by this ladder, and report which level you used in "time_precision":
   - "stated": an article or official source explicitly states the event time. Rationale must cite WHICH source (title or URL), QUOTE the stated time with its timezone, and SHOW the conversion to UTC (e.g., "NASA blog: 'scrubbed at 12:00 p.m. EST Feb 3' -> 17:00 UTC").
   - "convention": no explicit time, but the event has a documented standard release time (e.g., "BLS Employment Situation releases at 8:30 a.m. ET -> 12:30 UTC"). Name the convention in the rationale.
   - "date_only": only a date is determinable. Do NOT invent an hour: use T00:00:00Z of the earliest date the evidence was public, and say so in the rationale.
   Never present a guessed hour as stated.

Return JSON only -- an array with EXACTLY one entry per crux listed in the input:
[
  {"model": "<model name as given>",
   "event_group": "<short label, e.g. 'wdr-scrub' or 'deadline'>",
   "rule": "yes_event" | "no_deadline" | "no_irreversible",
   "adjudicated_datetime": "YYYY-MM-DDTHH:MM:SSZ",
   "time_precision": "stated" | "convention" | "date_only",
   "changed": true|false,
   "rationale": "1-3 sentences; for event-timestamped rules, cite the source and show the timezone conversion to UTC"}
]
\end{promptbox}



\end{document}